\documentclass[11pt]{article}

\usepackage[final]{acl}

\usepackage{times}
\usepackage{latexsym}

\usepackage[T1]{fontenc}

\usepackage{microtype}

\usepackage{inconsolata}

\usepackage{graphicx}

\usepackage{multirow}
\usepackage{tabularx}
\usepackage{booktabs}
\usepackage[most]{tcolorbox}
\usepackage{enumitem}
\usepackage{siunitx}
\usepackage{adjustbox}
\usepackage{needspace}

\newcommand{\promptplaceholder}[1]{\texttt{\textless #1\textgreater}}

\newtcolorbox{promptbox}[1]{
  colback=yellow!10!white,
  colframe=black,
  colbacktitle=yellow!50!white,
  coltitle=black,
  rounded corners,
  boxrule=0.8mm,
  width=\textwidth,
  fonttitle=\large,
  title={\textbf{#1}},
  left=2mm,
  right=2mm,
  top=2mm,
  bottom=2mm
}

\usepackage{xurl}
\newcolumntype{Y}{>{\raggedright\arraybackslash}X}
\newcolumntype{P}[1]{>{\raggedright\arraybackslash}p{#1}}
\newcommand{\checkpoint}[1]{\path{#1}}

\title{LLMs Learn Better In-Context from Rules than from Examples}

\author{
 \textbf{Xiang Fu\textsuperscript{1,*}}\quad
 \textbf{Seungmin Cho\textsuperscript{1,*}}\quad
 \textbf{Yukyung Lee\textsuperscript{1}}\quad
 \textbf{Najoung Kim\textsuperscript{1}}
\\
 \textsuperscript{1}Boston University
\\
{
{\texttt{\{xfu, minjo, ylee5, najoung\}@bu.edu}}
}}

\begin{document}
\maketitle

\begingroup
\renewcommand\thefootnote{}\footnotetext{
\textsuperscript{*}Equal contribution.} 

\endgroup

\begin{abstract}
Large language models (LLMs) exhibit in-context learning capabilities, where they can learn new tasks from prompt contexts without weight updates. We compare the learning efficacies of two prominent modes of in-context learning: (1) learning from descriptions of \textit{rules} (instruction following); and (2) learning from \textit{examples} of input-output demonstrations (few-shot prompting). Through five learning tasks that cover diverse domains (games, arithmetic, linguistic inferences), we compare two modes of learning (rules vs. examples) specifying the same underlying task. We furthermore explore model and task properties that modulate the learning efficacies. We find that models generally learn more reliably from rules than from examples alone, and additional examples on top of rules or simply scaling up the number of examples do not lead to consistent and significant gains. Instruction tuning amplifies the benefit of rule-based learning while keeping example-based learning capacities intact. Surprisingly, we find no privileged effect of example-based learning in base models, and rules still lead to gains in algebraic task domains. Overall, the comparative efficacy of rules over examples is larger when the task recruits algebraic abstractions and computations, and smaller when the task requires distributional sensitivity and/or recruits parametric knowledge.

\noindent\centering
\begin{adjustbox}{width=\linewidth}
\begin{tabular}{@{}c@{\hspace{6pt}}l@{}}
  \raisebox{-0.25\height}{\includegraphics[height=11pt]{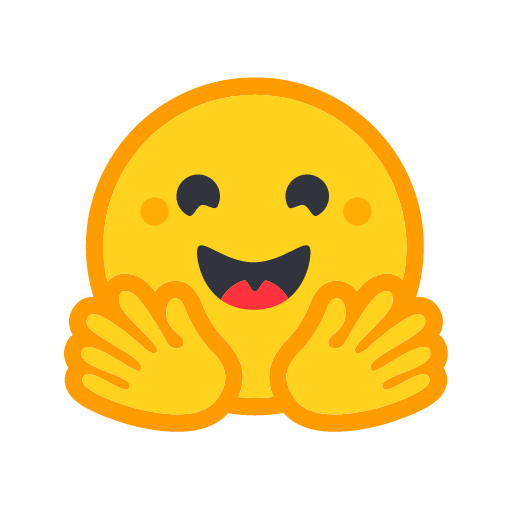}} &
  \href{https://huggingface.co/datasets/tin-lab/rules_vs_examples}{\texttt{datasets/tin-lab/rules\_vs\_examples}} \\
  \raisebox{-0.25\height}{\includegraphics[height=11pt]{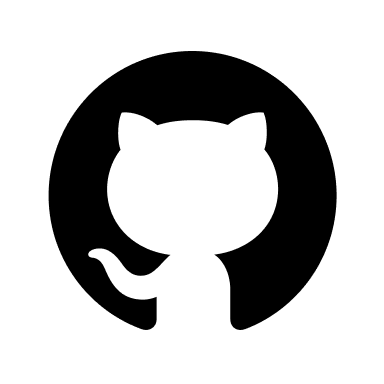}} &
  \href{https://github.com/tinlaboratory/rules-vs-examples}{\texttt{tinlaboratory/rules-vs-examples}} \\
\end{tabular}
\end{adjustbox}
\end{abstract}

\begin{figure*}[t]
  \includegraphics[width=\textwidth]{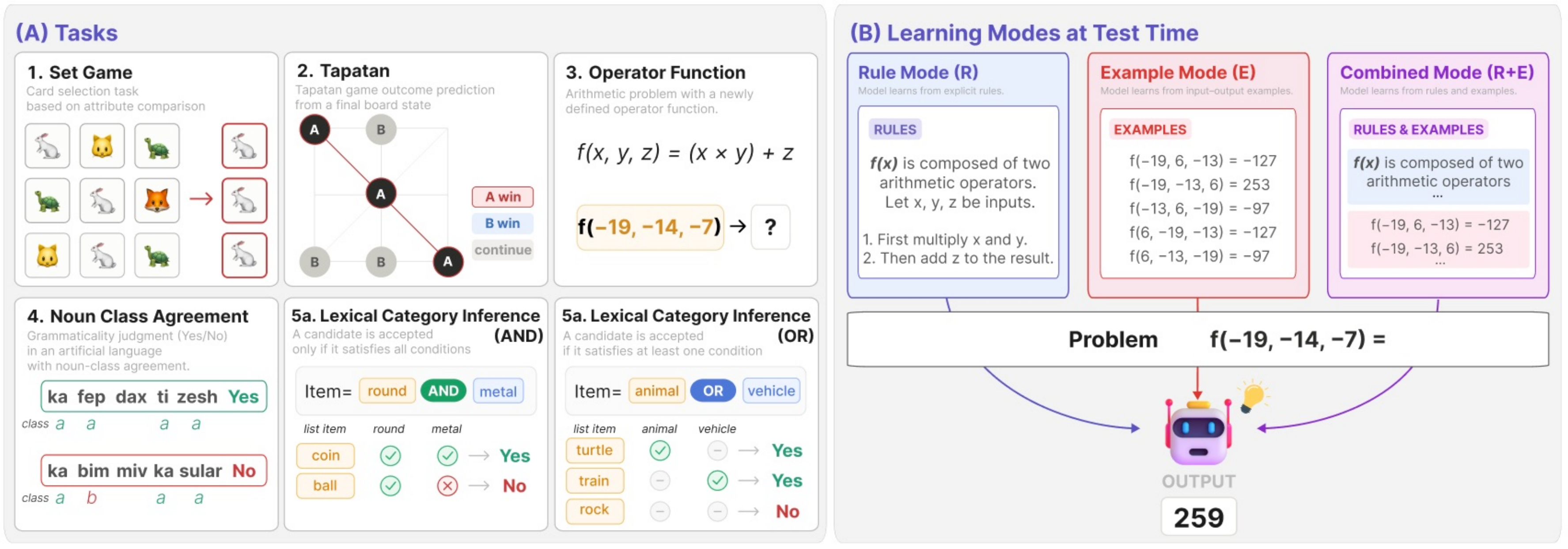}
  \caption{Visualization of our experiment design. (A) illustrates our tasks and (B) shows how the same task is presented to the model in different learning conditions.}
  \label{fig:task-design}
\end{figure*}

\section{Introduction}

Large language models (LLMs) have been shown capable of in-context learning, where a task can be learned from prompt contexts without weight updates \citep{Brown+2020FewShot,chung2024scaling,Lampinen+2024Spectrum}. Two predominant ways in which the task is communicated in context are: (1) through verbal descriptions or instructions for the task \citep{sanh2022multitask,chung2024scaling} (instruction- or \textit{rule}-based learning); and (2) through input-output demonstrations \citep{Brown+2020FewShot} (demonstration- or \textit{example}-based learning).

Recent work \citep{Davidson+2025TaskRepresentation} showed that even when both types of learning specify the same underlying task, LLMs do not induce a single common function vector representation, but instead activate partly overlapping mechanisms. They take this as support for the practice of combining rules and examples for the same task. However, distinct mechanisms do not necessarily indicate additive performance benefits when they are combined. In this work, we aim to conduct a systematic comparison of rule- vs. example-based learning of novel complex tasks to investigate their learning efficacies, and furthermore elucidate the properties of the learners and the tasks that modulate learning efficacy. The most similar work to ours is \citet{Liu+2024IncompleteLoop}, who reported as a part of their finding that rule-based learning generally outperforms example-based when the given rules are correct, although their main focus was relations to rule inference.

We compare rule- vs. example-based learning through in-context learning experiments with a suite of novel tasks. We focus on tasks that require complex reasoning, meaning that they involve inferring more than a single relational property or a single latent variable that underlies valid input-output mappings (e.g., the country-capital task). Our tasks cover domains of games (Set Game and Tapatan), arithmetic (Operator Function), and linguistic inferences (Noun Class Agreement and Lexical Category Inference). We instantiate each task in three different learning setups: rules, examples, and combined (rules + examples), and evaluate them on the same test items. This design lets us hold the latent task fixed while varying only how the task is communicated.

\Needspace{10\baselineskip}
Our research questions are as follows:
\begin{itemize}[itemsep=0pt]
    \item RQ1: Do models learn complex, novel tasks more reliably from rules (instructions) or from examples (demonstrations)?
    \item RQ2: What is the effect of the learner on the learning efficacy from rules vs. examples, especially the effect of instruction tuning?
    \item RQ3: What are the properties of the task itself that lends itself to better/worse learning from rules or examples?
\end{itemize}

\noindent Across task domains and model types, we find that learning from rules overall outperforms learning from (minimum coverage) examples. Providing examples on top of rules or scaling up the number of in-context examples do not lead to consistent and significant gains. Furthermore, the degree to which rules provide an advantage varies significantly depending on the nature of the task, where the comparative efficacy of rules over examples is larger when the task recruits algebraic abstractions and computations, and smaller when the task requires distributional sensitivity and/or recruits parametric knowledge. This opens up interesting future work for \textit{predicting} the efficacy of different in-context learning approaches based on task specifications, as well as better methods for tasks where rule-based learning yields limited gains.

\section{Method}
We design five tasks covering diverse domains, and instantiate these tasks in rule- and example-based in-context prompts. We additionally include a ``combined'' prompt, where both rules and examples are given. Each task domain consists of three task variants, corresponding to three difficulty levels (discussed further in Section~\ref{subsec:difficulty}). We programmatically generate a dataset for each task and split them into training and test sets; the training set is only relevant when there are examples given in context. All learning setups evaluate on the same set of test items. We provide more details about the experiments in Appendix~\ref{sec:appendix-experiment-setting} and \ref{sec:appendix-dataset-setting}.

\subsection{Models}
We evaluate both base and instruction-tuned checkpoints from Gemma 3, Qwen 2.5, Qwen 3, and OLMo 3/3.1, with model sizes ranging from 7B to 32B parameters. We also evaluate GPT-5.4 \citep{openai2026gpt54} through the OpenAI API, using the API model identifier \texttt{gpt-5.4}, as a strong reference point. The complete model list and compute setup are shown in Appendix~\ref{app:model-list}, Table~\ref{tab:model-list}.

\subsection{Learning Conditions}

Each task is presented to the target model in the following three learning conditions. See Table \ref{tab:prompt-structure}, Appendix~\ref{sec:appendix-rules-prompts} for all prompt structures.

\paragraph{Rules-only.}
The prompt contains a verbal description of the task rules followed by the input of the test problem. The rule description explicitly states the valid answer format. The rule prompts were revised through several iterations to ensure that the intended tasks are conveyed faithfully and to remove unintended ambiguity. We did not optimize the prompts for task performance. Full rule prompts can be found in Appendix~\ref{sec:appendix-rules-prompts}.

\paragraph{Examples-only.}
The prompt contains valid input-output demonstrations followed by the input of the test problem. No explicit task description or formatting instruction is provided, but every demonstration directly exposes the expected output format. The demonstration set furthermore provides all output labels (where applicable) as per the minimum coverage design (Section~\ref{subsec:example-selection}). Example input-output mappings for each task are shown in Appendix~\ref{subsec:appendix-dataset-construction}.

\paragraph{Combined.}
The prompt contains the same rule description as the rules-only condition, followed by input-output examples from the examples-only condition, and then the test problem. The combined condition therefore provides both types of information.

\subsection{Task Difficulty Conditions}
\label{subsec:difficulty}
We create three versions of each task corresponding to three difficulty levels: Easy, Medium, and Hard. These levels are associated with the number of variables that must be represented and tracked, the possible values of those variables, and the number of inference steps needed to compute the output. The advantage of rule-based task descriptions is that they constrain the hypothesis space more tightly compared to a fixed set of examples, leading to more robust coverage of possible input-outputs including corner cases. This benefit may become clearer as the space of possible input-output mappings grows (i.e., the difficulty level as we operationalize it increases). We implement the difficulty conditions to test whether in-context learning in LLMs does indeed exhibit the in-principle benefits of rule-based learning. If this is the case, we would observe more robust learning outcomes over difficulty levels in rule-based learning compared to example-based learning. We briefly discuss how difficulty is operationalized in the description of each task in Section~\ref{subsec:task-design}, and provide a full description in Table~\ref{tab_task_complexity}, Appendix~\ref{app:task-difficulty}.

\begin{figure}[t]
  \includegraphics[width=\columnwidth]{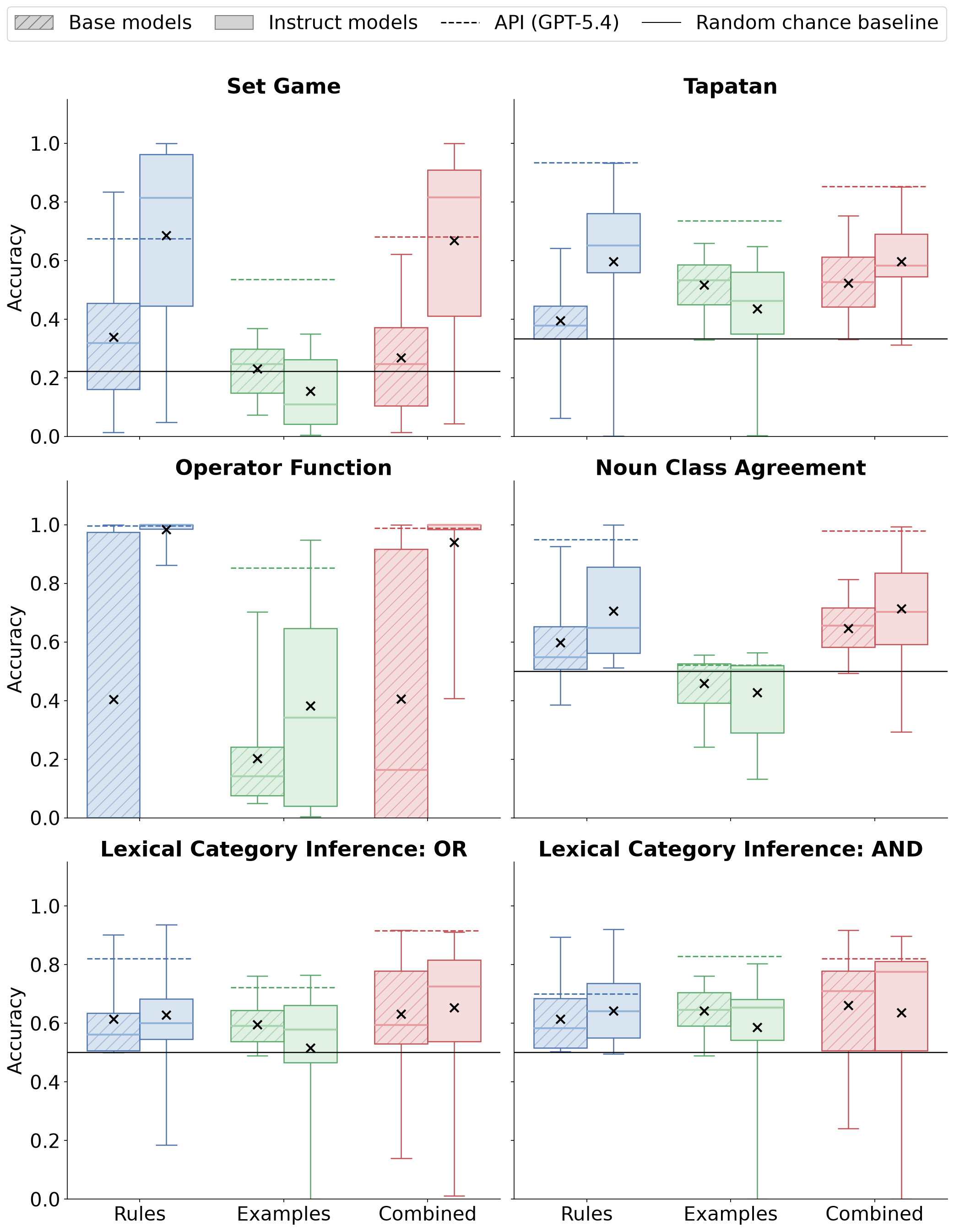}
  \caption{Accuracy by task, mode, and difficulty level, with base and instruction-tuned models shown separately. Hatched boxplots (left in each pair) denote base models, and solid boxplots (right) denote instruction-tuned models. The x markers denote mean accuracy, and horizontal dashed lines represent the performance of GPT-5.4 for comparison.  Random chance baselines (black lines) are provided for tasks with finite answer space where random chance can be calculated.}
  \label{fig:rule-example-box}
\end{figure}

\subsection{Task Design}
\label{subsec:task-design}

Our five tasks cover diverse domains including games, arithmetic, and linguistic inferences. We discuss the design of each task below. Although some of the tasks are inspired by known tasks (e.g., Set Game), we expect them to be sufficiently novel learning problems due to the variation we introduce in their specification (e.g., the set rule may require only two cards having the same attribute value, and the other to be different, which deviates from the original rules of the game). Furthermore, it has been shown empirically that such variations do make these tasks non-trivial to learn in context \citep{Wu+2023Counterfactual}. For each task, we also briefly discuss our hypotheses about the effects of the learning conditions based on the nature of the tasks---the task set is designed to cover a diverse range of hypotheses, where our priors vary regarding whether rule- or example-based learning will be effective for the task.

\subsubsection{Set Game}
This task is a variation of set game from \citet{Wu+2023Counterfactual} (Figure~\ref{fig:task-design}, A1). A player needs to select 3 cards out of 9 cards on the board that form a ``set''. Each card is associated with a range of attributes (e.g., animal, biome), and whether three cards form a set is determined by predefined rules that make reference to those attributes. Given the information about the 9 cards on the board, the model must select three cards that form a set according to the (explicit or latently inferrable) rules that determine a valid set. We predict that rule-based learning would be more effective for this task because same-or-different judgments relevant for set decisions require algebraic abstractions over variables, which may be more difficult for models to infer distributionally over examples. We expect the difficulty of this task to be modulated by the number of attributes, number of values each attribute can have, and the complexity of the rules that determine the conditions for a set.

\subsubsection{Tapatan}
This task is a variation of Tapatan, a two-player board game (Figure~\ref{fig:task-design}, A2). Tapatan is traditionally played on a $3 \times 3$ board, where two players take turns to place their pieces on a board until all pieces are placed, and then the pieces are moved around to adjacent coordinate points. The goal of the game is to place three pieces in a row before the other player. Horizontal, vertical, or diagonal rows are all considered valid. For this task, the model is provided with a board state rendered as an \(n \times n\) grid. Each cell contains \texttt{A}, \texttt{B}, or \texttt{.}, indicating a piece from Player \(A\), a piece from Player \(B\), or an empty position. Rows are shown from top to bottom and columns from left to right; the cell in column \(x\) and row \(y\) has coordinate \((x,y)\).

Given that there are two players (A and B), the possible game outcomes are: Player A wins, Player B wins, continue (i.e., no player has won). Solving this task correctly requires recognizing the win conditions determined by adjacency of the coordinates. Since adjacency is easy to recognize as patterns in the board state depiction, we expect example-based learning to be effective for this task, although we do not have a hypothesis about its efficacy compared to rules. We expect the difficulty of this task to be modulated by the size of the board ($n \times n$), the number of pieces each player can place, and the number of pieces that is considered a winning row.

\begin{figure}[t]
  \includegraphics[width=\columnwidth]{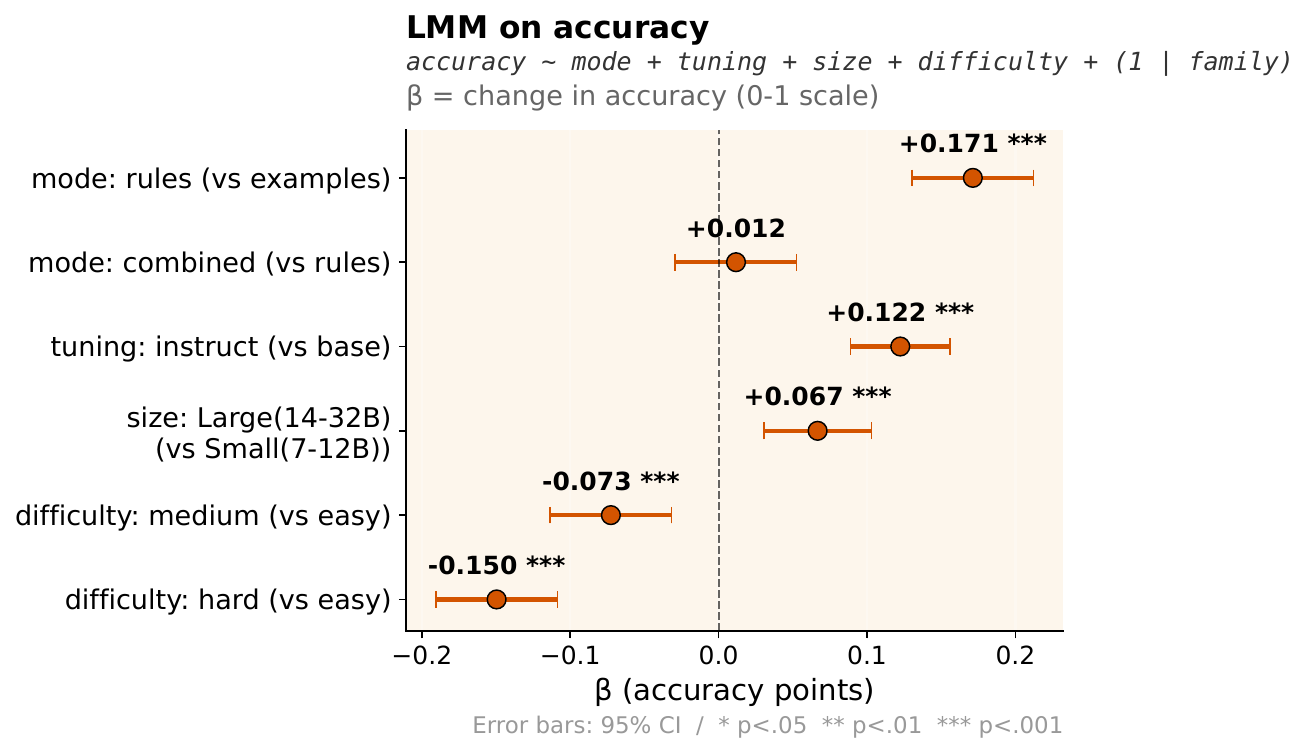}
  \caption{Main LMM effects of learning condition, instruction tuning, model size, and difficulty on accuracy.}
  \label{fig:rule-example-main}
\end{figure}

\subsubsection{Operator Function}
Operator Function (Figure~\ref{fig:task-design}, A3) is designed to evaluate models' learning efficacy in the arithmetic domain. Operator Function introduces a new arithmetic operator composed of familiar primitive operators (e.g., multiplication, addition). For this task, the model is given an arithmetic problem with a newly defined operator that is composed of \(n\) operators with \(n+1\) arguments, and must answer the problem using the newly defined operator function. We predict that rule-based learning will be much more effective for this task since the task requires algebraic reasoning and sequential application of operations. We expect the difficulty of the task to be modulated by the number of primitive operators that constitute the function (which also linearly increases the number of arguments, since all of our primitive operators are binary).

\subsubsection{Noun Class Agreement}
Noun Class Agreement (Figure~\ref{fig:task-design}, A4) is an artificial language learning task where the key information that needs to be learned is the class of the nouns in the language. The target languages exhibit agreement phenomena where forms of words in other syntactic categories (e.g., adjectives, verbs) change depending on the class of the noun they agree with. The form of the task is grammaticality judgment, where the model must output Yes or No depending on whether the given sentence is grammatical. The ungrammaticality of the examples solely derive from noun class agreement violations. Since noun class agreement is something that is typically learned distributionally during language acquisition, we predict that example-based learning would be effective (although we do not have a hypothesis about the learning efficacy compared to rule-based learning) and the task would likely benefit from a large number of examples. We expect the difficulty of the task to be modulated by the number of noun classes and the number of agreement phenomena in the target language.

\begin{figure*}[t]
  \includegraphics[width=0.5\textwidth]{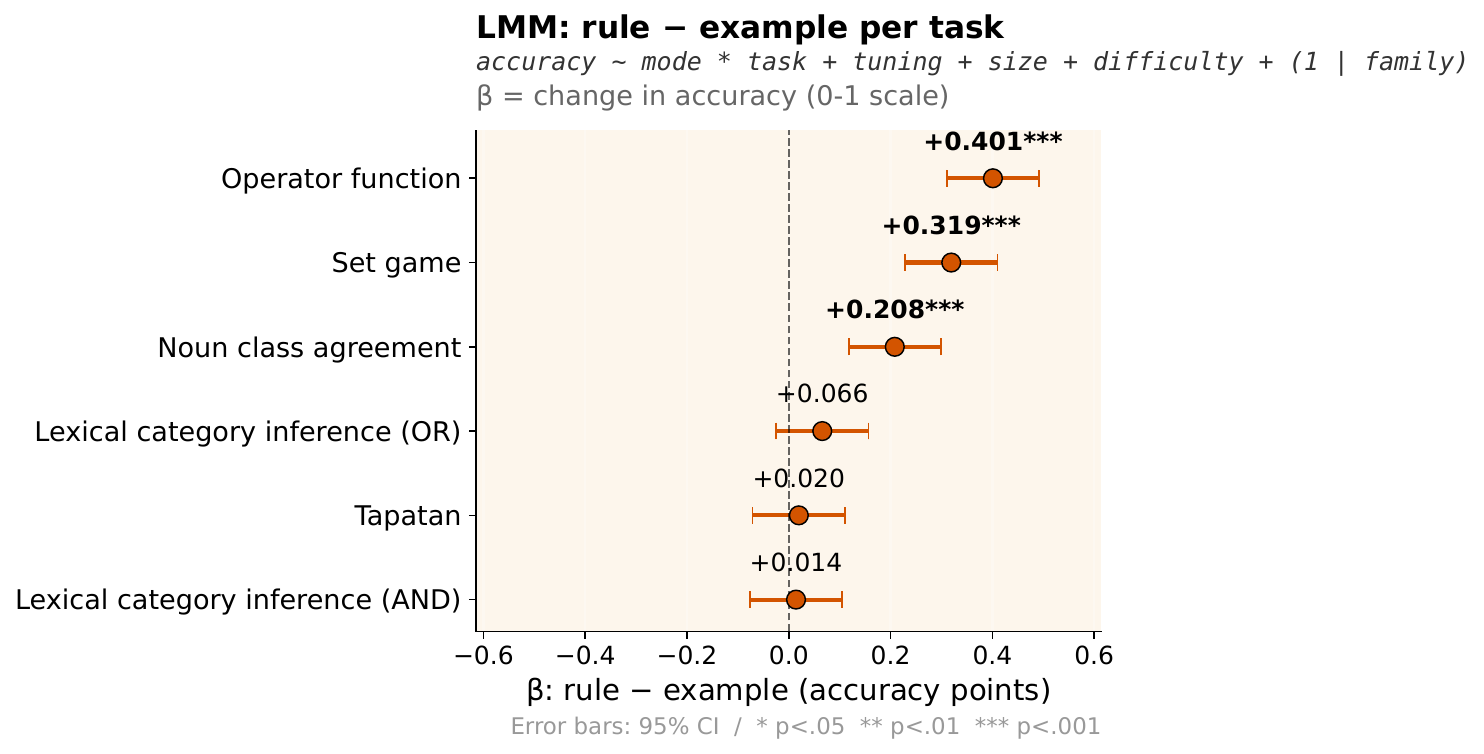}
  \hfill
  \includegraphics[width=0.5\textwidth]{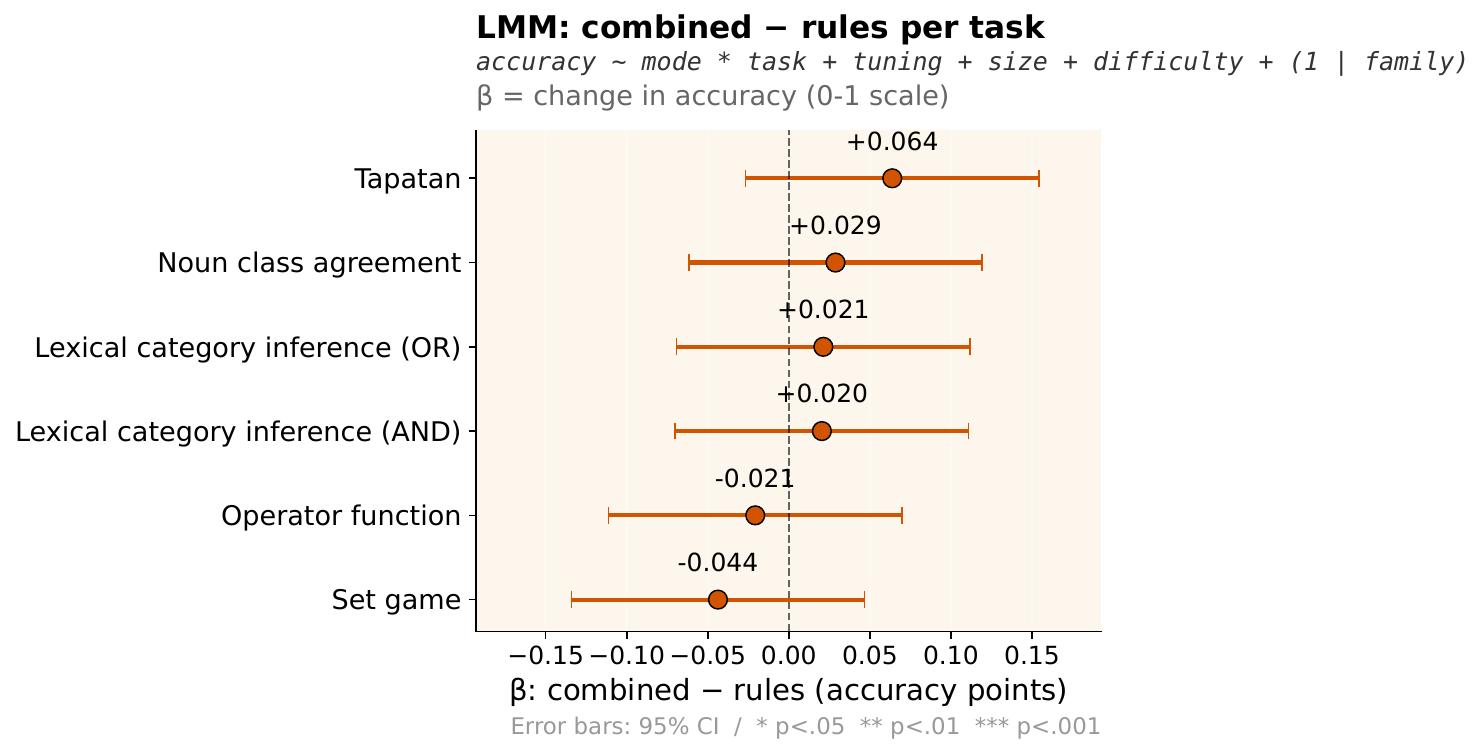}
  \caption{Task-specific LMM contrasts for rules minus examples (left) and combined minus rules (right). Panel scales differ, so distances should be compared only within panels.}
  \label{fig:rule-example-combined-task-mod}
\end{figure*}

\subsubsection{Lexical Category Inference}
\label{subsubsec:lexical-category-inference}
Lexical Category Inference (Figure~\ref{fig:task-design}, A5a \& A5b) task is also a sequence judgment task, but the judgment depends on semantic category membership rather than grammaticality. This task is specifically designed to recruit parametric knowledge to evaluate its effect on in-context learning efficacy. The task presents a sequence of four words and asks whether the sequence is valid, where a model must answer either Yes or No. Each word position in the sequence is associated with a lexical category, and a sequence is considered valid if each word in the sequence is a member of the specified lexical category. The category definitions are semantic in nature (e.g., birds, kitchen items, lives in South America). The category definitions could be made more complex by conjunctions (AND) or disjunctions (OR) of atomic category definitions (e.g., birds AND lives in South America, birds OR lives in South America). We include both variants to test the hypothesis that example-based learning, which likely relies on distributional cues, will perform well on conjunctive but not disjunctive category definitions. Disjunctive semantic category definitions are likely harder to induce from examples because positive examples are not distributionally similar; indeed, disjunctive category inference has been shown to be more difficult than conjunctive in human adults and children \citep{bruner1956study,snow1969conjunctive}. On the other hand, since rules explicitly surface the logical operators, we hypothesize the divergence between AND and OR to be smaller for rule-based learning. Finally, we expect the difficulty of the task to be modulated by the number of conjoined or disjoined category definitions.

\subsection{Example Selection}
\label{subsec:example-selection}

Examples-only and combined learning conditions require ``training examples'': i.e., input-output demonstrations that are shown in context. For all tasks other than Lexical Category Inference, datapoints from the same sampling process are divided randomly into training and test splits. A single example for Lexical Category Inference poses a unique category inference problem, so a single example consists of a set of training and test items. These sets of examples are randomly divided into training and test splits. The in-context demonstration sets used in our main experiments are minimum coverage, meaning that they include all possible output labels where applicable, all task primitives, and representative cases. The minimum coverage criterion is intended to ensure that there is enough information to infer the underlying task, but we do not optimize demonstration selection further. Still, our demonstrations are likely more systematic than typical user-created few-shot prompts, which often contain only a small number of heuristically selected examples. For each test item, we sample the demonstration set (with the minimum coverage constraint) and randomly order the demonstrations, rather than using one fixed set and ordering across all test items. This prevents the results being dependent on the idiosyncrasies of a particular set of demonstrations and ordering. Different tasks and difficulty levels accordingly use different numbers of examples. The task-specific minimum coverage criteria and example counts are reported in Appendix~\ref{app:example-coverage} and Table~\ref{tab:example-coverage}. We additionally examine the effect of scaling the size of the demonstration set in Appendix~\ref{sec:appendix-example-scaling}.

\section{Results}

\subsection{Effect of Learning Conditions}

\paragraph{Rules vs Examples.}
The results are shown in Figure~\ref{fig:rule-example-box} (blue vs. green boxplots). GPT-5.4 is displayed only as a reference point and is excluded from our linear mixed model (LMM) analyses; all LMMs are fit only to the open-weight models. The open-weight instruction-tuned models generally showed patterns similar to GPT-5.4, suggesting that the open-weight results may generalize to stronger models. We find that rule-based learning significantly outperforms example-based learning overall (Figure~\ref{fig:rule-example-main}, \texttt{mode: rules (vs examples)}). However, this benefit of rule-based learning was not observed universally across all tasks. Figure~\ref{fig:rule-example-combined-task-mod} (left) shows that rules are the most beneficial for Set Game and Operator Function tasks, corroborating our predictions for these tasks. On the other hand, benefits for Tapatan (where the board representation provides strong distributional cues) and the linguistic tasks (Noun Class Agreement and Lexical Category Inference) are either smaller in magnitude or not significant. Within the linguistic tasks, rules do not yield clear benefits for the more semantic task that recruits parametric knowledge (Lexical Category Inference).

Furthermore, the conjunctive versus disjunctive contrast in Lexical Category Inference supports our prediction about example-based learning (results in Appendix~\ref{sec:appendix-lexical-logic}). Under example-based learning, conjunctive category definitions yield higher accuracy than disjunctive definitions, while the same pattern does not consistently appear under rule-based or combined learning. We discuss this contrast and its interaction with difficulty further in Appendix~\ref{sec:appendix-lexical-logic}.

\paragraph{Combined learning.}
Combined learning closely tracks the pattern of rule-based learning, as can be seen in Figure~\ref{fig:rule-example-box} (blue vs. red boxplots). Surprisingly, the additional examples in the combined condition does not provide statistically significant benefits over just rules (Figure~\ref{fig:rule-example-main}, \texttt{mode: combined (vs rules)}). This marginal effect of additional examples to rules is shared across all tasks, as shown in the per-task analysis in Figure~\ref{fig:rule-example-combined-task-mod} (right).

\begin{figure}[t]
\centering
  \includegraphics[width=0.95\columnwidth]{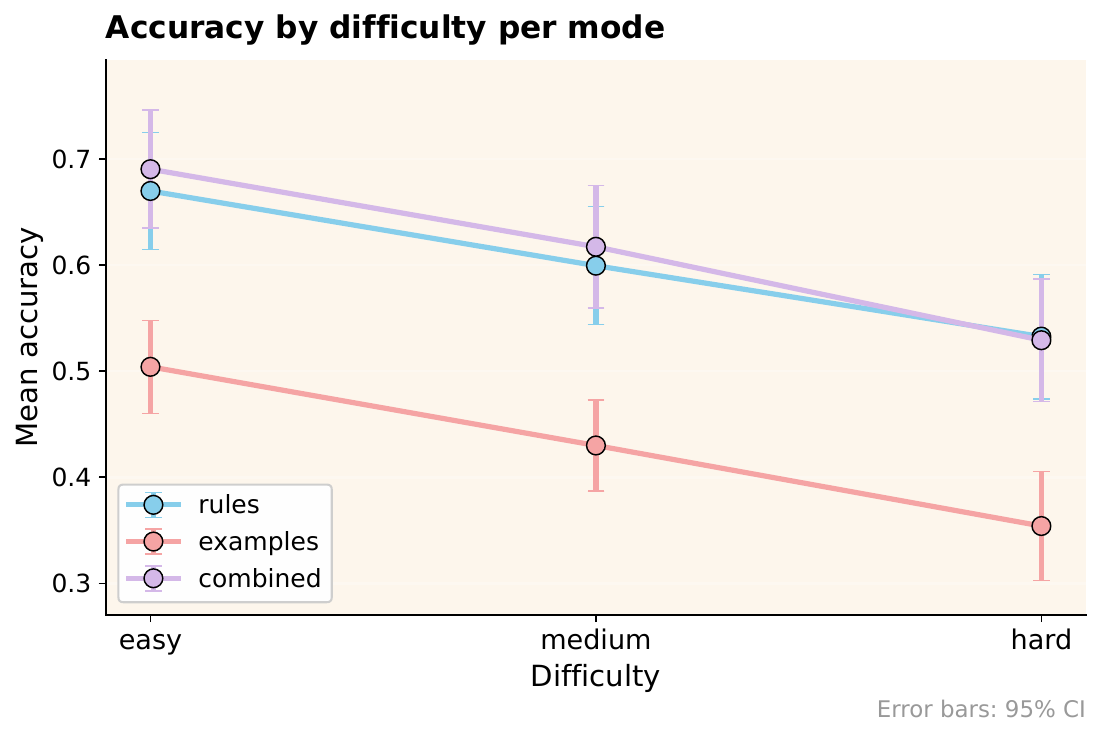}
  \caption{Slope of decline for increasing task difficulty.}
  \label{fig:difficulty}
\end{figure}

\paragraph{Problem difficulty.}
As Figures~\ref{fig:rule-example-main} (\texttt{difficulty}) and \ref{fig:difficulty} show, manipulating the task parameters that we hypothesized to control difficulty in Section~\ref{subsec:task-design} does lead to monotonic performance drops across the three difficulty levels. However, the slopes of decline for the three learning conditions are almost parallel, showing that rules are not particularly advantageous over examples in terms of \textit{robustness} across difficulty levels, contra our hypothesis in Section~\ref{subsec:difficulty}.

\paragraph{Number of examples.} As discussed in Section~\ref{subsec:example-selection}, the set of examples we used in our main experiments were minimum coverage. To examine the effect of larger numbers of examples which have been shown to improve in-context learning \citep{Agarwal+2024ManyShot}, we conduct experiments scaling the number of examples with models in the OLMo family (Appendix~\ref{sec:appendix-example-scaling}). We observe that there is substantial task and model variation in the example scaling trends. One common observation is that not many task/model combinations lead to monotonic gains over rule-based learning---often the gains are flat, show diminishing returns, or even yield drops in performance. This suggests that in modern LLMs, rule-based learning is a very effective mode of in-context learning that remains comparable to, and often better than, even a large number of in-context examples. We furthermore observe that when examples do yield benefits, smaller base models tend to benefit more, and substantial gains are typically only observed for easier versions of the tasks. These results complement findings from \citet{Agarwal+2024ManyShot} and \citet{bertsch-etal-2025-context}, contributing scenarios under which scaling the number of in-context examples does not lead to gains or even lead to performance degradation. Although we leave detailed investigations for future work, one possible hypothesis for this discrepancy is our minimum coverage demonstration selection, which limits gains from improved coverage by scaling.

\subsection{Effect of Model Properties}

\begin{figure}[t!]
\centering
  \includegraphics[width=0.95\columnwidth]{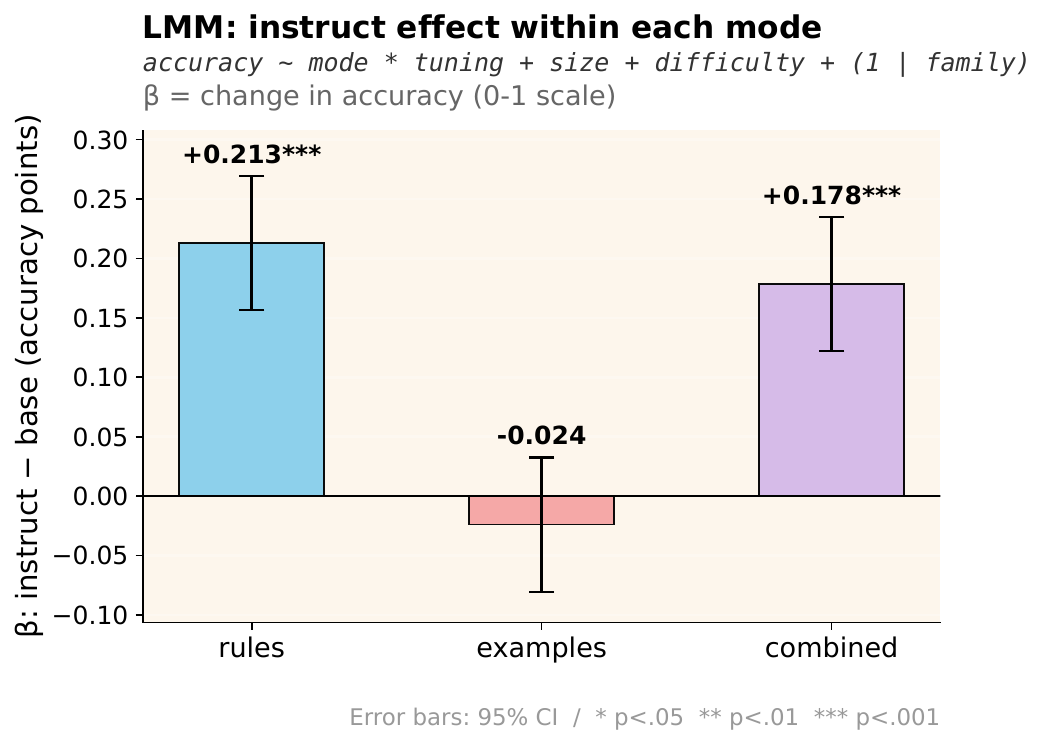}
  \caption{Instruction-tuning effects within each learning condition.}
  \label{fig:tuning_effect_mode}
\end{figure}

\paragraph{Base vs. instruction-tuned models.}
Instruction tuning significantly improves performance for rule-based and combined learning, but does not significantly affect example-based learning (Figure~\ref{fig:tuning_effect_mode}). This suggests that instruction-tuned models are overall superior in-context learners to base models, keeping the example-based learning capacity intact while improving upon rule-based learning. The gains in rule-based learning are likely more than an improved ability to follow the correct output format in the absence of demonstrations, since combined learning overall has little benefit over rules in base models (Appendix~\ref{sec:appendix-additional-stats}, Figure~\ref{fig:by-task-cr-tuning}).

\begin{figure}[t!]
  \centering
  \includegraphics[width=0.95\columnwidth]{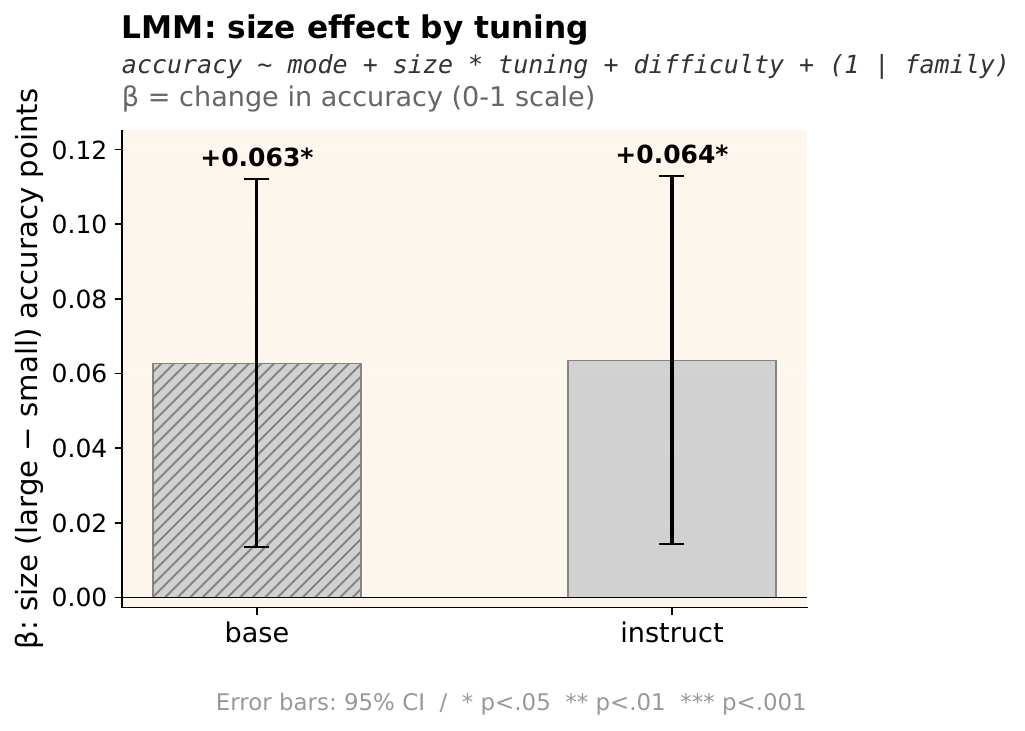}
  \caption{Model-size effects for base and instruction-tuned models.}
  \label{fig:model-size}
\end{figure}

\paragraph{Model size.}
We analyze the effect of model size on learning efficacy by grouping the models we tested into smaller (7B--12B) and larger (14B--32B). Figure \ref{fig:model-size} shows that there are significant gains for both base and instruction-tuned models with increasing model size.

\section{Discussion}
\label{sec:discussion}

\paragraph{RQ1: Models learn complex, novel tasks more reliably from rules, and examples on top of rules do not lead to additional gains.} We found that overall, rules were the more effective medium of learning new tasks in context. Nevertheless, the degree to which rules are more effective than examples was modulated by the nature of the individual tasks, which we will discuss further in the context of RQ3. To our surprise, providing examples on top of rules did not lead to statistically significant gains in general, suggesting that in modern LLMs, rules yield robust enough task representations to make the accompanying correct demonstrations redundant. This overall lack of benefit from additional examples may be a reflection of our revision process of the rule prompts for completeness and clarity, since the reported gains from examples in the literature are focused on settings where rules are likely underspecified (e.g., \citealt{wang-etal-2022-super}).

\paragraph{RQ2: Instruction tuning improves rule-based learning and does not degrade example-based learning. Example-based learning is not privileged in base models.}
Unsurprisingly, instruction tuning improved rule-based learning substantially. Two more notable observations are that (1) instruction tuning does not degrade the capacity to learn from examples and (2) example-based learning has no particular advantage in base models. As for the second observation, we did not observe better learning outcomes from examples in base models in general (Appendix~\ref{sec:appendix-additional-stats}, Figure~\ref{fig:by-task-re-tuning}), and the tasks that benefited the most from rules overall (Operator Function and Set Game) likewise showed statistically significant benefits of rule-based learning in base models. Rule-based learning explicitly surfaces information that must be latently inferred in example-based learning (e.g., the task, the label space, and the expected output format: \citealt{Min+2022RethinkingDemonstrations,Pan+2023TaskRecognition}). Therefore, the difficulty of the inference at hand may be in general easier in rule-based learning, insofar as the learner possesses the capacity to understand and follow rules. This gap is also reflected in our statement of the task-specific hypotheses: we hypothesized certain tasks to lend itself better to example-based learning, but were agnostic about whether they will be \textit{more} effective than rules. But as stated previously, to make use of information provided in the rules, the learner must already have the capacity to do so, which is not a trivial precondition. The comparative efficacy of rule-based learning in both base and instruct models indicates that the capacity to follow verbal rules emerges even without instruction tuning, although instruction tuning does amplify it.

\paragraph{RQ3: Recruitment of parametric knowledge and distributional sensitivity makes rule-based learning less effective.}
As discussed in the answer to RQ1, the benefit of rule-based learning over examples did not uniformly hold across all tasks. Then, what are the properties of tasks that make rule-based learning less effective? Overall, our game and arithmetic tasks (Operator Function and Set Game) showed more rule-superiority than linguistic tasks (Noun Class Agreement and Lexical Category Inference). We interpret this as the effect of tasks that require algebraic reasoning vs. (only) category membership inference. Even within category membership inference tasks, category definitions that recruit parametric knowledge (Lexical Category Inference; categories are semantically defined) versus category definitions that are purely formal (Noun Class Agreement; categories are arbitrary) show different trends, where the former benefits less from rules. The conjunctive versus disjunctive comparison further refines this pattern (Appendix~\ref{sec:appendix-lexical-logic}), showing that the rule advantage is much stronger for disjunctive category definitions and less clear for conjunctive category definitions, where similarity-based inference is expected to be more effective for conjunctive than disjunctive.

\paragraph{Future work}
We have identified several model- and task-level factors that modulate the learning efficacies from rules vs. examples. While our experiments focused on diverse coverage of task domains to broadly explore possible factors, future work could further investigate the properties we identified, conducting more controlled hypothesis testing (e.g., variations of a single task to modulate parametric knowledge recruitment). More concrete findings from these controlled studies could motivate work that lets us predict in-context learning efficacies based on task specifications without running the experiment, as well as methods to improve learning. Mechanistic analyses as in \citet{Davidson+2025TaskRepresentation} could also be conducted to test whether findings about function vectors for simple tasks like country-capital analogies generalize to more realistic, complex tasks like ours.

\section{Related Work}

\paragraph{Different modes of in-context learning.}
In-context learning first became prominent under the setting we call example-based learning in this work, where few-shot examples are given in the prompt \citep{Brown+2020FewShot}. Follow-up studies have shown that the performance of example-based learning is sensitive to many different factors, including example selection, retrieval, ordering, prompt design, and the number of examples \citep{Rubin+2021RetrievePrompts,lu-etal-2022-fantastically,sclar2024quantifying,Agarwal+2024ManyShot}. Another prominent in-context learning setting that we referred to as rule-based learning provides a set of rules or instructions in natural language, again given to the model as a part of the prompt \citep{sanh2022multitask,chung2024scaling}. \citet{Lampinen+2024Spectrum} surveyed a wider range of in-context learning setups and proposes a more generalized, broader-coverage definition of in-context learning. The most closely related work to ours is \citet{Liu+2024IncompleteLoop}, who compared instruction following (rules), few-shot prompting (examples), and instruction inference for the same tasks, although their rule prompts included examples. Their work shares one converging conclusion to ours that rules outperform examples when the rules are correct, although the work mainly focuses on the relations of these learning setups with instruction inference. Our work focuses on a more detailed investigation into the comparison between rule- and example-based learning, relating them to model/task properties.

\paragraph{Task representations.}
Recent work has also asked whether different in-context learning conditions elicit common task representations. \citet{Davidson+2025TaskRepresentation} found that rule- and example-based learning conditions recruit partially overlapping but distinct mechanisms. While they took this as support for combining rules and examples, our results show that there are no consistent additive benefits for the combined learning condition in terms of task performance. This is not at odds with their finding since distinct mechanisms do not necessarily indicate additive performance benefits when combined. Furthermore, our finding that instruction tuning improves rule-based learning while leaving example-based learning intact is consistent with the two-mechanism view. Although only marginally discussed in the paper, \citet{Davidson+2025TaskRepresentation} also reported behavioral results corroborating our findings where rule-based learning of much simpler tasks yields better performance than 10-shot example-based learning in smaller models ($\leq$ 8B). Concurrent work by \citet{Yang+2026LexicalTaskHeads} identified lexical task representations that are shared across rule and example prompts. Our work is largely behavioral, and given that our tasks are an order of magnitude more complex than the tasks used in these studies, our tasks could be interesting future targets for mechanistic analyses.

\paragraph{Characterizations of rule following and induction in LLMs.}
Work on rule induction and concept learning has shown that abstractions recovered from examples can support downstream reasoning, and also that in-context concept learning can be shaped by simplicity and task structure \citep{Zhu+2023LearnRules,Wang+2024BooleanComplexity}. Recent work furthermore has shown that LLMs still struggle with explicit constraints, compositional rule systems, and rules that conflict with pretrained regularities \citep{Mu+2024SimpleRules,Wang+2024ULogic,Wu+2023Counterfactual}, which may limit the efficacy of rule-based learning.

\paragraph{In-context vs. in-weights learning.}
While our work primarily investigated efficacies of different in-context learning setups, prior work also has compared properties of learning in-context vs. in-weights. For instance, \citet{chan2022transformers} argued that generalization from in-weights and in-context information shows different trends. \citet{Cook+2026ProgrammingBackprop} compared in-weights instructions to in-context instructions for executing procedural tasks, finding that in-weights instructions are less reliable.

\paragraph{Rule- and example-based learning in humans.}
Cognitive science research has long been interested in rule- vs. exemplar-based learning and generalization \citep{nosofsky1986attention,smith1994similarity,tenenbaum1999rules,gentner1998similarity,Dasgupta+2022RulesExemplars}, especially for categorization problems. However, few studies have directly compared human learning efficacy when rules and examples specify the same underlying task. The most relevant work can be found in education research where learning from direct instructions and worked examples are analyzed for pedagogical efficacy \citep{SwellerCooper+1985WorkedExamples,Atkinson+2000WorkedExamples,Kang+2022IBLReview}, although not compared against each other. Studies of grammar and second-language learning show that instruction, exposure, and rule search can support different learning outcomes \citep{DeKeyser+1995LearningGrammarRules,Doughty+1991SLInstruction,Robinson+1996SLRules}. Work on individual differences further suggests that learners vary in their reliance on abstract rules versus memorized examples, leading to different learning outcomes \citep{McDaniel+2014StableTendencies,LittleMcDaniel+2015Memorization,Herzog+2025StableDifferences}. While we find a general disconnect between research on in-context learning literature in LLMs and human sciences, future work could explore their interactions further.

\section{Conclusion}
In this work, we compared two prominent modes of in-context learning: from natural language descriptions of rules and from examples of valid input-output mappings, through controlled learning experiments using novel, complex tasks spanning a wide range of domains. By evaluating open-weight LLMs from three different model families, we find that rules are the more effective medium of in-context learning, and additional examples on top of rules do not yield statistically significant performance gains. While the efficacy of rules (over examples) is overall greater in instruction-tuned models, base models followed a similar trend, with no privileged efficacy of example-based learning over rules. More detailed analyses of individual task properties suggest that the gap between rule- and example-based learning is larger when the task recruits algebraic abstractions and computations, and smaller when the task requires distributional sensitivity and/or recruits parametric knowledge.

\section*{Limitations}
Our experiments are designed to isolate how models use rules, examples, and their combination when the same latent task is presented through different prompt formats. This design makes the comparison systematic, but it may narrow the generalizability of the conclusions to real-world applications. Set Game and Tapatan are variations of existing games, so pretraining exposure to their original forms may influence performance. Although our specifications differ from the original games and counterfactual designs are known to present challenging learning problems \citep{Wu+2023Counterfactual}, we did not audit model pretraining data and therefore cannot fully guarantee the counterfactual manipulations' novelty. Our rule prompts were revised for clarity and disambiguation without performance-based prompt search or selection. This controlled prompt construction process does not reflect many practical settings, where user instructions may be ambiguous, incomplete, underspecified, noisy, or expressed through informal language. Our results therefore should not be read as showing that rules are always preferable modes of task specification in practical applications. Rather, they indicate that when a faithful rule description can be supplied, current models often use it more reliably than examples alone. Future work should investigate whether the current rule advantage persists when rules and examples are less idealized, which would better establish its practical applicability. All experiments in this work are academic explorations using controlled synthetic tasks, and we do not foresee any direct risks associated with this research.

\section*{Code and Data Availability}
Code and experiment materials are available at
\url{https://github.com/tinlaboratory/rules-vs-examples}.
The generated data used in our experiments are available on Hugging Face at
\url{https://huggingface.co/datasets/tin-lab/rules_vs_examples}.

\section*{Acknowledgments}
XF was supported by Boston University's Undergraduate Research Opportunities Program (UROP). We acknowledge that the computational work reported in this paper was performed on the Shared Computing Cluster, which is administered by \href{https://www.bu.edu/tech/support/research/}{Boston University's Research Computing Services}.

\bibliography{custom}

\newpage

\appendix

\section{Experimental Setup}
\label{sec:appendix-experiment-setting}

\subsection{Task Difficulty}
\label{app:task-difficulty}

Table \ref{tab_task_complexity} reports the number of primitives and inference steps for each task and difficulty level.

\begin{table*}[p]
  \centering
  \small
  \setlength{\tabcolsep}{4pt}
  \renewcommand{\arraystretch}{1.18}
  \begin{tabular}{@{}P{0.13\textwidth}P{0.08\textwidth}P{0.37\textwidth}P{0.37\textwidth}@{}}
    \toprule
    \textbf{Task} & \textbf{Level} & \textbf{Primitives} & \textbf{Inference steps} \\
    \midrule

    Set game
      & Easy
      & 3 cards; 2 attributes per card; 10 kinds per attribute.
      & Check 2 attributes; each attribute must be all the same or all different. \\

      & Medium
      & 3 cards; 3 attributes per card; 10 kinds per attribute.
      & Check 3 attributes; each attribute must be all same or all different. \\

      & Hard
      & 3 cards; 3 standard attributes per card; 1 number attribute per card; 10 kinds per standard attribute; numbers from 1 to 12.
      & Check 3 same-or-different attributes; check number condition; numbers must have two same attributes and one different. \\

    \addlinespace[4pt]
    \midrule
    \addlinespace[4pt]

    Tapatan
      & Easy
      & \(3{\times}3\) board; up to 3 pieces per player; 3-in-a-line win condition.
      & Parse the final board grid; map row and column positions to coordinates; check whether either player has a valid horizontal, vertical, or diagonal 3-in-a-line row under the parity-sensitive adjacency rule; output \(A\) win, \(B\) win, or continue. \\

      & Medium
      & \(5{\times}5\) board; up to 4 pieces per player; 4-in-a-line win condition.
      & Parse the final board grid; map row and column positions to coordinates; check whether either player has a valid horizontal, vertical, or diagonal 4-in-a-line row under the parity-sensitive adjacency rule; output \(A\) win, \(B\) win, or continue. \\

      & Hard
      & \(7{\times}7\) board; up to 5 pieces per player; 5-in-a-line win condition.
      & Parse the final board grid; map row and column positions to coordinates; check whether either player has a valid horizontal, vertical, or diagonal 5-in-a-line row under the parity-sensitive adjacency rule; output \(A\) win, \(B\) win, or continue. \\

    \addlinespace[4pt]
    \midrule
    \addlinespace[4pt]

    Operator function
      & Easy
      & 3 arguments; 2 operators.
      & Apply first operation; carry intermediate result; apply second operation. \\

      & Medium
      & 4 arguments; 3 operators.
      & Apply 3 ordered operations; carry each intermediate result forward. \\

      & Hard
      & 5 arguments; 4 operators.
      & Apply 4 ordered operations; carry the longest chain of intermediate results. \\

    \addlinespace[4pt]
    \midrule
    \addlinespace[4pt]

    Noun class agreement
      & Easy
      & 2 noun classes; 8 nouns; 2 determiners; 2 noun phrases per sentence.
      & Check determiner agreement; check both noun phrases. \\

      & Medium
      & 4 noun classes; 8 nouns; 4 determiners; 4 adjective suffixes; 2 noun phrases per sentence.
      & Check determiner agreement; check adjective-suffix agreement; check both noun phrases. \\

      & Hard
      & 6 noun classes; 12 nouns; 6 determiners; 6 adjective suffixes; 6 subject prefixes; 6 object suffixes; 2 noun phrases and 1 marked verb per sentence.
      & Check determiner agreement; check adjective-suffix agreement; check verb subject-prefix agreement; check verb object-suffix agreement. \\

    \addlinespace[4pt]
    \midrule
    \addlinespace[4pt]

    Lexical category inference
      & Easy
      & 4 category positions; 4 single-condition category definitions; 4-word candidate sequence.
      & Check word-category membership; check category order; decide full-sequence validity. \\

      & Medium
      & 4 category positions; 4 two-condition category definitions; conjunctive or disjunctive relation; 4-word candidate sequence.
      & Evaluate 2 conditions per category; apply AND or OR relation; check category order. \\

      & Hard
      & 4 category positions; 4 three-condition category definitions; conjunctive or disjunctive relation; 4-word candidate sequence.
      & Evaluate 3 conditions per category; apply AND or OR relation; check category order. \\

    \bottomrule
  \end{tabular}
  \caption{\label{tab_task_complexity}Difficulty manipulations for our five tasks. For each task, the table summarizes how the easy, medium, and hard levels change the primitives that must be represented and the inference steps needed to determine the answer.}
\end{table*}

\subsection{Modeling Details}
\label{app:model-list}

We evaluate base and instruction-tuned checkpoints from the Gemma 3, OLMo 3/3.1, and Qwen families, referring to family-level model reports for documentation \citep{GemmaTeam2025Gemma3,TeamOlmo2025Olmo3,QwenTeam2024Qwen25,QwenTeam2025Qwen3}. For the 32B OLMo models, we use \checkpoint{allenai/Olmo-3-1125-32B} as the base checkpoint from the OLMo 3 release and \checkpoint{allenai/Olmo-3.1-32B-Instruct} as the instruction-tuned checkpoint from the subsequent OLMo 3.1 release. This difference in release names only reflects the checkpoint names made available by Ai2 (i.e., not a different model family under our experimental setup). We also evaluate GPT-5.4 through the OpenAI API \citep{openai2026gpt54}, using the API model identifier \texttt{gpt-5.4}. GPT-5.4 is reported only as a separate reference point and is excluded from our statistical analyses, all of which are fit only to the open-weight checkpoints. Table~\ref{tab:model-list} reports the exact checkpoint and API identifiers and inference hardware used in our experiments.

\begin{table*}[!t]
  \centering
  \small
  \setlength{\tabcolsep}{4pt}
  \renewcommand{\arraystretch}{1.14}

  \newcommand{\modelgroup}[2]{%
    \textbf{#1}\\[-1pt]
    {\scriptsize #2}%
  }

  \begin{tabularx}{\textwidth}{@{}P{0.16\textwidth}P{0.055\textwidth}Y Y P{0.075\textwidth}@{}}
    \toprule
    \textbf{Family / Access Terms}
      & \textbf{Size}
      & \textbf{Base Model}
      & \textbf{Instruction-tuned Model}
      & \textbf{Hardware} \\
    \midrule

    \multirow{2}{0.16\textwidth}{\modelgroup{Gemma 3}{Gemma Terms of Use}}
      & 12B
      & \checkpoint{google/gemma-3-12b-pt}\newline
      {\scriptsize Context limit: \num{131072} tokens}
      & \checkpoint{google/gemma-3-12b-it}\newline
      {\scriptsize Context limit: \num{131072} tokens}
      & NVIDIA L40S \\
      & 27B
      & \checkpoint{google/gemma-3-27b-pt}\newline
      {\scriptsize Context limit: \num{131072} tokens}
      & \checkpoint{google/gemma-3-27b-it}\newline
      {\scriptsize Context limit: \num{131072} tokens}
      & NVIDIA RTX A6000 \\

    \addlinespace[4pt]

    \multirow{2}{0.16\textwidth}{\modelgroup{OLMo 3/3.1}{Apache-2.0}}
      & 7B
      & \checkpoint{allenai/Olmo-3-1025-7B}\newline
      {\scriptsize Context limit: \num{65536} tokens}
      & \checkpoint{allenai/Olmo-3-7B-Instruct}\newline
      {\scriptsize Context limit: \num{65536} tokens}
      & NVIDIA L40S \\
      & 32B
      & \checkpoint{allenai/Olmo-3-1125-32B}\newline
      {\scriptsize Context limit: \num{65536} tokens}
      & \checkpoint{allenai/Olmo-3.1-32B-Instruct}\newline
      {\scriptsize Context limit: \num{65536} tokens}
      & NVIDIA RTX A6000 \\

    \addlinespace[4pt]

    \modelgroup{Qwen2.5}{Apache-2.0}
      & 32B
      & \checkpoint{Qwen/Qwen2.5-32B}\newline
      {\scriptsize Context limit: \num{131072} tokens}
      & \checkpoint{Qwen/Qwen2.5-32B-Instruct}\newline
      {\scriptsize Context limit: \num{32768} tokens}
      & NVIDIA RTX A6000 \\

    \addlinespace[4pt]

    \multirow{2}{0.16\textwidth}{\modelgroup{Qwen3}{Apache-2.0}}
      & 8B
      & \checkpoint{Qwen/Qwen3-8B-Base}\newline
      {\scriptsize Context limit: \num{32768} tokens}
      & \checkpoint{Qwen/Qwen3-8B}\newline
      {\scriptsize Context limit: \num{40960} tokens}
      & NVIDIA L40S \\
      & 14B
      & \checkpoint{Qwen/Qwen3-14B-Base}\newline
      {\scriptsize Context limit: \num{32768} tokens}
      & \checkpoint{Qwen/Qwen3-14B}\newline
      {\scriptsize Context limit: \num{40960} tokens}
      & NVIDIA RTX A6000 \\

    \addlinespace[4pt]

    \modelgroup{OpenAI}{OpenAI API terms}
      & N/A
      & N/A
      & \checkpoint{gpt-5.4}\newline
      {\scriptsize Context limit: \num{1050000} tokens}
      & API \\

    \bottomrule
  \end{tabularx}

  \caption{Models, inference hardware, context limits, and license or access terms used in the experiments. We report exact Hugging Face checkpoint identifiers for the open-weight models and the OpenAI API model identifier. Each model entry also reports its context limit. For open-weight checkpoints, this is the configured maximum total sequence length; for GPT-5.4, it is the documented API context window. The OLMo 32B row pairs the OLMo 3 base checkpoint with the instruction-tuned checkpoint released subsequently under the OLMo 3.1 name. The Qwen3 instruction-tuned checkpoints are hybrid-thinking models; thinking mode was disabled using \texttt{enable\_thinking=False}, as detailed below.}
  \label{tab:model-list}
\end{table*}

\paragraph{Model licenses and access terms.}
The Gemma 3 checkpoints are governed by Google's Gemma Terms of Use \citep{GoogleGemmaTerms2026}. The OLMo 3 and OLMo 3.1 checkpoints are released under Apache-2.0 \citep{Ai2Olmo3ModelCards2026}. The Qwen2.5-32B and Qwen3 checkpoints used in our experiments are released under Apache-2.0 \citep{QwenTeam2024Qwen25License,QwenTeam2025Qwen3Repo}. GPT-5.4 is not an open-weight model; it was accessed only through the OpenAI API and is governed by OpenAI's API and service terms \citep{OpenAIServiceAgreement2025}. We did not redistribute model weights.

\paragraph{Compute budget.}
All experiments were inference-only; no model training, fine-tuning, or weight updates were performed. We estimate the local inference budget at approximately 600 GPU-hours, summed over evaluation jobs run on L40S and RTX A6000 GPUs. Table~\ref{tab:model-list} reports the inference hardware used for each open-weight checkpoint. This estimate includes the main experiments' rules-only, examples-only, and combined evaluations for all open-weight models, the OLMo example-scaling experiments, and reruns needed for failed or incomplete inference jobs. We compute GPU-hours as the number of GPUs allocated to an evaluation job multiplied by its wall-clock duration, summed across completed runs. GPT-5.4 was evaluated through the OpenAI API and is not included in the local GPU-hour total.

\paragraph{Inference and decoding settings.}
All models were evaluated with a shared decoding policy across tasks and learning conditions. Generation length caps were specified separately, as described below. For open-weight models, we used Hugging Face Transformers \citep{Wolf+2020Transformers} for checkpoint and tokenizer utilities, including chat-template handling, and vLLM-based serving \citep{Kwon+2023vLLM} for inference. We set \texttt{dtype="auto"} in vLLM; for all 14 open-weight checkpoints, this resolved to \texttt{bfloat16}, consistent with the dtype specified in each checkpoint's model configuration. This includes all six checkpoints evaluated on NVIDIA L40S GPUs and all eight checkpoints evaluated on NVIDIA RTX A6000 GPUs. Both GPU types support BF16, so the hardware split did not introduce FP16-versus-BF16 precision variation. No vLLM quantization was enabled unless explicitly specified. Table~\ref{tab:model-list} reports the configured context limit for every open-weight checkpoint and the documented API context window for GPT-5.4. These limits apply to the total sequence length, including prompt and generated tokens.

Base models were prompted as text-completion models using the raw task prompt strings. For each instruction-tuned open-weight checkpoint, the complete task prompt was wrapped as a single user message and rendered using the checkpoint tokenizer's default chat template through \texttt{tokenizer.apply\_chat\_template}, with \texttt{tokenize=False} and \texttt{add\_generation\_prompt=True}; the resulting rendered prompt was then passed to vLLM. For the hybrid-thinking Qwen3 instruction-tuned checkpoints, \texttt{Qwen/Qwen3-8B} and \texttt{Qwen/Qwen3-14B}, the same call additionally set \texttt{enable\_thinking=False}. This hard switch disabled the generation of \texttt{\textless think\textgreater...\textless/think\textgreater} content. We did not use the prompt-level \texttt{/no\_think} instruction or a vLLM reasoning parser. GPT-5.4 was queried through an OpenAI-compatible chat API endpoint using the API model identifier \texttt{gpt-5.4}, with the complete task prompt supplied as a single user message.

Generation used deterministic decoding for all reported evaluations, with sampling disabled by setting temperature to \(0.0\). Maximum generation length was specified as a cap on new tokens. For base models, we used short caps (between \(2\) and \(50\) new tokens) since all tasks had constrained answer formats: \(2\) to \(8\) tokens for Lexical Category Inference, \(4\) for Tapatan, \(8\) for Noun Class Agreement, and \(50\) for Set Game and Operator Function. For instruction-tuned models, which more often produced explanatory text around the answer, we used larger task- or model-specific caps when needed to avoid premature truncation. For example, Operator Function used a cap of \(6{,}000\) new tokens, and Lexical Category Inference used fixed model-specific caps, commonly starting at \(8{,}192\) new tokens. These caps were fixed before full evaluation and held constant within each model and task setting across rules-only, examples-only, and combined learning conditions.

\section{Dataset Details}
\label{sec:appendix-dataset-setting}

\paragraph{Dataset generation and sampling.}

Dataset generation used a base seed of \(42\). We split the dataset into \(n\) training and \(m\) test examples and held these splits fixed for all evaluations. We used seed \(123\) for demonstration selection and ordering for all tasks. For each test item, a separate demonstration set was sampled from the training split under the task-specific minimum-coverage criteria described in Appendix~\ref{app:example-coverage}. The ordering of the demonstration items was also randomized. The reported results therefore aggregate over many randomized demonstration sets and orderings for the same task, rather than using a fixed demonstration set and ordering for all examples. The fixed seed is used to reproduce these assignments and does not imply that a single global demonstration set or ordering was used for a task and difficulty setting. The same test items and demonstration configurations were reused across checkpoints and matched between the examples-only and combined conditions, enabling paired comparisons.

\subsection{Dataset Construction}
\label{subsec:appendix-dataset-construction}

\paragraph{Set Game.}
Each example presents a board containing nine cards paired with one valid set of three cards. Cards are generated from combinations of the task-specific card attributes, and each board is created by randomly sampling distinct cards. The generator evaluates every three-card combination on the board and retains boards for which at least one valid set exists. Boards that duplicate an earlier sampled board are rejected. Candidate boards are also downsampled to reduce redundancy among easily solvable boards and to maintain variation in the resulting board configurations.

\begin{figure}[h]
\centering
\begin{tcolorbox}[title=Set Game, width=\columnwidth]
\small\ttfamily\raggedright\sloppy

\textbf{Input}\\

        ["axolotl","swamp"],
        ["axolotl","tundra"],
        ["bat", "badland"],
        ["bat","hill"],
        ["bat","plain"],
        ["bat","tundra"],
        ["rabbit","plain"],
        ["rabbit","swamp"],
        ["sheep","hill"]

\textbf{Output}\\

["bat","hill"], ["bat","plain"], ["bat","tundra"]

\end{tcolorbox}
\caption{An input-output example for Set Game.}
\label{fig:set-game-demo}
\end{figure}

\paragraph{Tapatan.}
Each example is a board state together with its outcome under the Tapatan win condition. Difficulty changes the board size, the maximum possible number of pieces per player, and the required winning-line length. The generator derives board states from legal Tapatan games, but the model is shown only the final board configuration, not the move sequence that led to it. The board is rendered as an \(n \times n\) grid whose cells contain \texttt{A}, \texttt{B}, or \texttt{.}, indicating a piece from Player \(A\), a piece from Player \(B\), or an empty position. The train and test splits are balanced over \(A\) win, \(B\) win, and continue labels.

\begin{figure}[h]
\centering
\begin{tcolorbox}[title=Tapatan, width=\columnwidth]
\small\ttfamily\raggedright\sloppy

\textbf{Input}\\
. B .\\
. . B\\
A A A

\textbf{Output}\\
A win

\end{tcolorbox}
\caption{An input-output example for Tapatan.}
\label{fig:tapatan-demo}
\end{figure}

\paragraph{Operator Function.}
Each example consists of a symbolic operator expression together with its evaluated output. Input expressions are generated by sampling nonzero integers from the range \([-20,20]\), with difficulty determining the number of arguments and composed operations. The hidden operator definition is then applied to compute the final output. The test set includes the operator expression and its answer. The demonstration set includes all argument-order permutations for the sampled inputs, yielding \(3!\), \(4!\), and \(5!\) demonstration variants for the easy, medium, and hard settings.

\begin{figure}[h]
\centering
\begin{tcolorbox}[title=Operator Function, width=\columnwidth]
\small\ttfamily\raggedright\sloppy

\textbf{Input}\\

[-10, 5, 10]

\textbf{Output}\\

-40

\end{tcolorbox}
\caption{An input-output example for Operator Function.}
\label{fig:operator-function-demo}
\end{figure}

\paragraph{Noun Class Agreement.}
Noun class agreement data are generated as matched pairs. Each pair contains one grammatical sentence and one minimally corrupted sentence derived from the same underlying structure. Difficulty changes the number of noun classes and the number of agreement sites that must be considered. Negative examples are created from a difficulty-specific set of violation types, including determiner mismatches, adjective-suffix mismatches, noun substitutions, and, at the hardest level, verb-prefix and verb-suffix mismatches.

\begin{figure}[h]
\centering
\begin{tcolorbox}[title=Noun Class Agreement, width=\columnwidth]
\small\ttfamily\raggedright\sloppy

\textbf{Input}\\

Sentence: ka sular dax ka wug

\textbf{Output}\\

Yes

\end{tcolorbox}
\caption{An input-output example for Noun Class Agreement.}
\label{fig:noun-class-agreement-demo}
\end{figure}

\paragraph{Lexical Category Inference.}
Each example contains four category definitions and one four-word candidate sequence. A positive example contains one valid word from each category in the required order. A negative example preserves the same surface format but violates the category specification; violations could include category order, category membership, or the logical relation defining a category. The easy setting uses single-condition categories. The medium and hard settings use two- and three-condition categories, respectively, and are generated separately for conjunctive and disjunctive definitions. In the conjunctive case, a word must satisfy all listed conditions for its category. In the disjunctive case, a word may satisfy any one of the listed conditions. The training and test sets are balanced over Yes and No answers, with negative cases selected to cover the main failure types.

\begin{figure}[h]
\centering
\begin{tcolorbox}[title=Lexical Category Inference, width=\columnwidth]
\small\ttfamily\raggedright\sloppy

\textbf{Input}\\
List: tuna, tangerine, dog, squirt gun $\rightarrow$ Yes \\
List: tuna, tangerine, dog, mackerel $\rightarrow$ No \\
List: duckling, dagger, monkey, spinach $\rightarrow$ Yes \\
List: crossbow, cherry, monkey, squirt gun $\rightarrow$ No \\
List: anchovy, sword, flashlight, ball $\rightarrow$ Yes \\
List: broccoli, pistol, streetlight, herring $\rightarrow$ No \\
List: trout, peach, bear, lego

\textbf{Output}\\
Yes

\end{tcolorbox}
\caption{An input-output example for Lexical Category Inference.}
\label{fig:lexical-category-inference-demo}
\end{figure}

\subsection{Label and category distributions}
\label{app:label-category-distributions}

Table~\ref{tab:appendix-split-distributions} summarizes the dataset statistics.

\begin{table*}[!t]
  \centering
  \footnotesize
  \setlength{\tabcolsep}{3pt}
  \renewcommand{\arraystretch}{1.12}
  \begin{tabularx}{\textwidth}{@{}P{0.15\textwidth}P{0.12\textwidth}Y Y  @{}}
    \toprule
    \textbf{Task} & \textbf{Setting} & \textbf{|Train|} & \textbf{|Test|} \\
    \midrule

    \multirow[t]{3}{0.13\textwidth}{Set game}
      & Easy
      & 3,000 (1000 for possible SET types)
      & 501 \\
      & Medium
      & 7,000 (1000 for possible SET types)
      & 504 \\
      & Hard
      & 7,000 (1000 for possible SET types)
      & 504 \\

    \addlinespace[3pt]\midrule\addlinespace[3pt]

    \multirow[t]{3}{0.13\textwidth}{Tapatan}
      & Easy
      & 6,000 (\(A\) win \(=2{,}000\), \(B\) win \(=2{,}000\), continue \(=2{,}000\))
      & 1,998 (\(A\) win \(=666\), \(B\) win \(=666\), continue \(=666\)) \\
      & Medium
      & 6,000 (\(A\) win \(=2{,}000\), \(B\) win \(=2{,}000\), continue \(=2{,}000\))
      & 1,998 (\(A\) win \(=666\), \(B\) win \(=666\), continue \(=666\)) \\
      & Hard
      & 6,000 (\(A\) win \(=2{,}000\), \(B\) win \(=2{,}000\), continue \(=2{,}000\))
      & 1,998 (\(A\) win \(=666\), \(B\) win \(=666\), continue \(=666\)) \\

    \addlinespace[3pt]\midrule\addlinespace[3pt]

    \multirow[t]{3}{0.13\textwidth}{Operator function}
      & Easy
      & 6,000 (1000 for each unique set of 3 integers)
      & 500 \\

      & Medium
      & 24,000 (1000 for each unique set of 4 integers)
      & 500 \\

      & Hard
      & 120,000 (1000 for each unique set of 5 integers)
      & 500 \\

    \addlinespace[3pt]\midrule\addlinespace[3pt]

    \multirow[t]{3}{0.13\textwidth}{Noun class agreement}
      & Easy
      & 6,000 (Yes \(=3{,}000\), No \(=3{,}000\))
      & 2,000 (Yes \(=1{,}000\), No \(=1{,}000\)) \\
      & Medium
      & 6,000 (Yes \(=3{,}000\), No \(=3{,}000\))
      & 2,000 (Yes \(=1{,}000\), No \(=1{,}000\)) \\
      & Hard
      & 6,000 (Yes \(=3{,}000\), No \(=3{,}000\))
      & 2,000 (Yes \(=1{,}000\), No \(=1{,}000\)) \\

    \addlinespace[3pt]\midrule\addlinespace[3pt]

    \multirow[t]{5}{0.13\textwidth}{Lexical category inference}
      & Easy
      & 6,000 (Yes \(=3{,}000\), No \(=3{,}000\))
      & 600 (Yes \(=300\), No \(=300\)) \\

      & Medium (OR)
      & 6,000 (Yes \(=3{,}000\), No \(=3{,}000\))
      & 600 (Yes \(=300\), No \(=300\)) \\

      & Hard (OR)
      & 6,000 (Yes \(=3{,}000\), No \(=3{,}000\))
      & 600 (Yes \(=300\), No \(=300\)) \\

      & Medium (AND)
      & 6,000 (Yes \(=3{,}000\), No \(=3{,}000\))
      & 600 (Yes \(=300\), No \(=300\)) \\

      & Hard (AND)
      & 6,000 (Yes \(=3{,}000\), No \(=3{,}000\))
      & 600 (Yes \(=300\), No \(=300\)) \\

    \bottomrule
  \end{tabularx}
  \caption{Dataset statistics. For the main experiments, examples were randomly sampled from the training set. See Table \ref{tab:example-coverage} for the number of examples used for each test run.}
  \label{tab:appendix-split-distributions}
\end{table*}

\subsection{Example Coverage}
\label{app:example-coverage}

In the examples-only and combined conditions, the number of in-context examples are determined by task-specific coverage criteria rather than by a shared fixed count, since the number of primitives and reasoning steps required vary by task and difficulty level (Table~\ref{tab_task_complexity}). For each task and difficulty, we construct a ``minimum coverage'' example set that covers all labels (if the task is classification), all primitives, and representative cases that are sufficient to infer the latent task from examples. Table~\ref{tab:example-coverage} summarizes these coverage targets and counts of the minimum coverage set. Our main experiments use these sets of examples, although we additionally report example scaling experiments in Appendix~\ref{sec:appendix-example-scaling}.

\begin{table*}[!t]
  \centering
  \footnotesize
  \setlength{\tabcolsep}{4pt}
  \renewcommand{\arraystretch}{1.12}
  \begin{tabularx}{\textwidth}{@{}P{0.13\textwidth}P{0.08\textwidth}cY@{}}
    \toprule
    \textbf{Task} & \textbf{Level} & \textbf{Count} & \textbf{Coverage target and rationale} \\
    \midrule

    \multirow[t]{3}{0.13\textwidth}{Set game}
      & Easy
      & 3
      & Covers targeted same-or-different configurations over two attributes. Each demonstration instantiates a relational pattern for deciding whether three cards form a valid set. \\
      & Medium
      & 7
      & Extends coverage to three attributes, including cases with one, two, or three shared attribute dimensions. \\
      & Hard
      & 7
      & Uses the medium attribute templates and adds the special number condition. The number attribute changes one rule dimension but does not introduce additional same-or-different templates. \\

    \addlinespace[3pt]\midrule\addlinespace[3pt]

    \multirow[t]{3}{0.13\textwidth}{Tapatan}
      & Easy
      & 12
      & Covers all three outcome labels and the major final-board configurations on the \(3{\times}3\) board. Four demonstrations per label keep \(A\) win, \(B\) win, and continue balanced while covering horizontal wins, vertical wins, diagonal wins, near misses, continue states, and different piece-count patterns. \\
      & Medium
      & 18
      & Preserves the same outcome and geometric coverage on the \(5{\times}5\) board. Six demonstrations per label add variation in 4-in-a-line winning configurations, diagonal parity cases, near misses, and non-winning final board states. \\
      & Hard
      & 24
      & Preserves label balance while adding coverage for sparser 5-in-a-line wins, longer valid lines, diagonal parity cases, near misses, continue states, and different occupancy patterns on the \(7{\times}7\) board. \\

    \addlinespace[3pt]\midrule\addlinespace[3pt]

    \multirow[t]{3}{0.13\textwidth}{Operator function}
      & Easy
      & 6
      & Covers all argument-order permutations for the three-argument function, allowing the model to observe how argument position affects the two-step computation. \\
      & Medium
      & 24
      & Covers all permutations of four arguments, exposing how each input position participates in the three-step composed function. \\
      & Hard
      & 120
      & Covers all permutations of five arguments for the longest and most order-sensitive composed function. \\

    \addlinespace[3pt]\midrule\addlinespace[3pt]

    \multirow[t]{3}{0.13\textwidth}{Noun class agreement}
      & Easy
      & 8
      & Covers the noun inventory and both determiner-mismatch positions. Positive examples expose the two noun classes, while negative examples isolate subject and object determiner violations under balanced Yes and No labels. \\
      & Medium
      & 12
      & Covers the four noun classes, noun-phrase agreement markers, and the six difficulty-specific violation types, including determiner, adjective-suffix, and noun-based errors. \\
      & Hard
      & 16
      & Covers the six-class inventory and the expanded agreement system, including determiner, adjective, verb-prefix, and verb-suffix errors. \\

    \addlinespace[3pt]\midrule\addlinespace[3pt]

    \multirow[t]{3}{0.13\textwidth}{Lexical category inference}
      & Easy
      & 24
      & Includes twelve positive and twelve negative demonstrations to expose the required category order, membership checks, and plausible negative candidate sequences. \\
      & Medium
      & 32
      & Covers two-condition conjunctive or disjunctive categories. The demonstrations balance Yes and No labels while showing valid component realizations and failures of order, membership, or logical relation. \\
      & Hard
      & 48
      & Expands component-level coverage for three-condition categories and includes near misses that preserve surface plausibility while violating the intended semantic boundary. \\

    \bottomrule
  \end{tabularx}
  \caption{Demonstration coverage conditions for example-based and combined learning.}
  \label{tab:example-coverage}
\end{table*}

\section{Example Scaling Experiments}
\label{sec:appendix-example-scaling}

\begin{figure*}[h!]
  \includegraphics[width=1\linewidth]{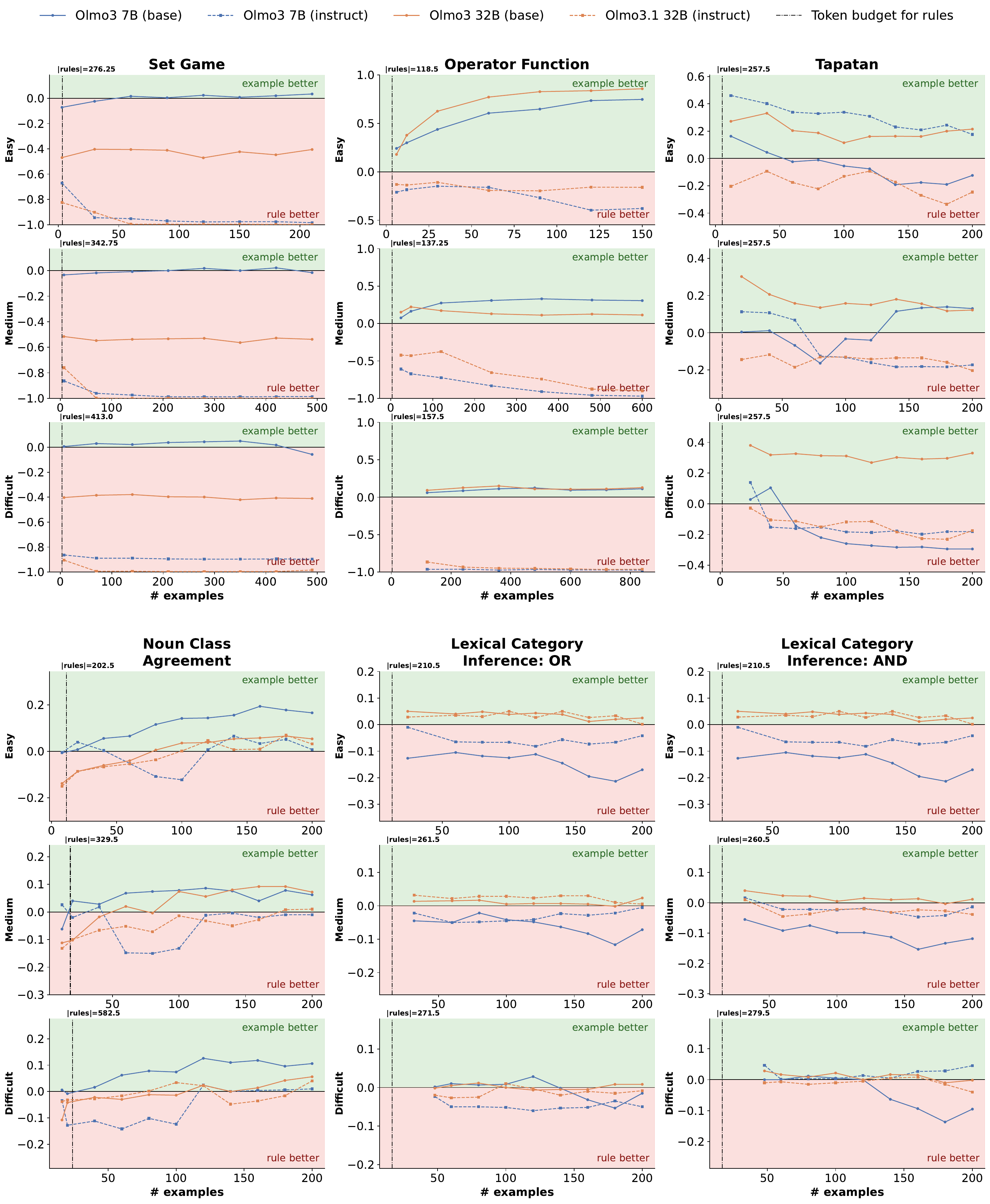}
  \caption{Example scaling experiment results for each task and difficulty level. The \(x\)-axis denotes the number of examples, and the \(y\)-axis denotes the accuracy difference between rule-based learning (fixed) and example-based learning, where the number of examples increases along the \(x\)-axis. The first \(x\) value in each graph corresponds to the minimum-coverage example set used in the main experiment. The green area corresponds to cases where example-based learning yields better learning outcomes than rule-based learning, and the red area corresponds to cases where rule-based learning yields better outcomes than example-based learning. The vertical dash-dot line indicates the point where the token budget of the example prompt is equivalent to the token budget of the rule prompt, where the average rule prompt token counts are annotated with \(|\mathrm{rules}|\). See Table~\ref{tab:token_budget} for the main experiment's token budget for each learning condition.}
  \label{fig:example-scaling}
\end{figure*}

Since the set of examples we use for the main experiments is minimum coverage, it is possible that scaling up the number of in-context examples would yield different outcomes. We conduct experiments with the four models in the OLMo 3 family (base/instruction-tuned $\times$ 7B/32B) to examine the effect of example scaling. Figure~\ref{fig:example-scaling} reports the effect of example scaling, measured as accuracy difference between rule-based learning (always fixed) and example-based learning (\# examples increases along the $x$-axis). The leftmost tick in every graph corresponds to the number of examples used in the main experiment. We find that whether example scaling leads to benefits over rules is highly task- and model-dependent, and example scaling can also often lead to no additional gains or even degradations in performance. When example scaling does help, gains most often appear at the Easy level, with most benefits to smaller base models. We note that many of the severe performance degradation scenarios had truncated responses (e.g., the drop to near-zero performance in instruct models for Set Game). In examining the outputs, we found that the additional examples led instruct models into loops of long, repetitive analysis of the patterns in the given examples, which was the main cause of premature truncation. Therefore, we expect that a longer truncation window is not likely to fix the observed performance degradation.

\begin{table*}[t]
  \centering
  \small
  \setlength{\tabcolsep}{7pt}
  \renewcommand{\arraystretch}{1.16}

  \begin{tabular*}{\textwidth}{@{\extracolsep{\fill}}
      P{0.25\textwidth}
      P{0.16\textwidth}
      S[table-format=4.2]
      S[table-format=4.2]
      S[table-format=4.2]
    @{}}
    \toprule
    \textbf{Task}
      & \textbf{Level}
      & {\textbf{Rules}}
      & {\textbf{Examples}}
      & {\textbf{Combined}} \\
    \midrule

    \multirow[t]{3}{0.25\textwidth}{\textbf{Set game}}
      & Easy   & 276.25 & 254.25  & 487.50  \\
      & Medium & 342.75 & 734.75  & 1006.50 \\
      & Hard   & 413.00 & 958.00  & 1280.00 \\

    \addlinespace[5pt]

    \multirow[t]{3}{0.25\textwidth}{\textbf{Tapatan}}
      & Easy   & 257.50 & 582.50  & 817.50  \\
      & Medium & 257.50 & 1645.50 & 1880.50 \\
      & Hard   & 257.50 & 3620.50 & 3855.50 \\

    \addlinespace[5pt]

    \multirow[t]{3}{0.25\textwidth}{\textbf{Operator function}}
      & Easy   & 118.50 & 206.00  & 365.00  \\
      & Medium & 137.25 & 1086.00 & 1213.00 \\
      & Hard   & 157.50 & 7022.00 & 7167.75 \\

    \addlinespace[5pt]

    \multirow[t]{3}{0.25\textwidth}{\textbf{Noun class agreement}}
      & Easy   & 202.50 & 142.50 & 322.50 \\
      & Medium & 329.50 & 261.50 & 568.50 \\
      & Hard   & 582.50 & 392.50 & 952.50 \\

    \addlinespace[5pt]

    \multirow[t]{5}{0.25\textwidth}{\textbf{Lexical category inference}}
      & Easy (Shared) & 210.50 & 410.50 & 630.50  \\
      & Medium (AND) & 260.50 & 543.50 & 813.50  \\
      & Medium (OR)  & 261.50 & 528.50 & 798.50  \\
      & Hard (AND)   & 279.50 & 782.50 & 1071.50 \\
      & Hard (OR)    & 271.50 & 781.50 & 1062.50 \\

    \bottomrule
    \end{tabular*}
    \caption{\label{tab:token_budget}
    Mean prompt token budgets for the default evaluation setting. Although the rule prompt text is fixed within each task and difficulty level, its tokenized length differs across checkpoints because base checkpoints receive raw text prompts, whereas instruction-tuned checkpoints include their default chat templates. Each value is averaged over the four OLMo checkpoints using the tokenizer and input formatting associated with each checkpoint. Rules, examples, and combined correspond to the three learning modes; lexical category inference reports each logic condition separately.}
\end{table*}

\section{Lexical Category Inference: Effect of Conjunctive (AND) vs. Disjunctive (OR) Category Definitions}
\label{sec:appendix-lexical-logic}

\begin{figure*}[t]
\centering
  \includegraphics[width=\linewidth]{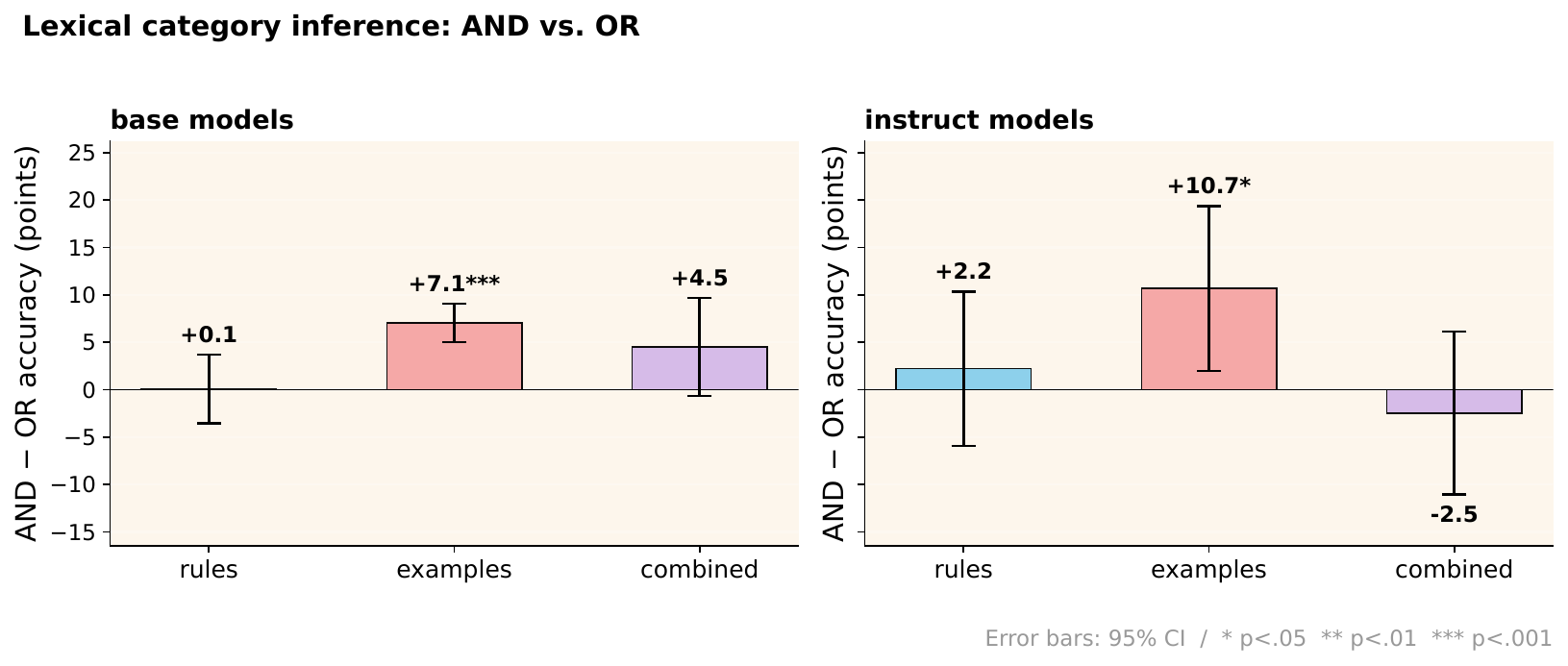}
  \caption{Estimated effect of conjunctive versus disjunctive category definitions within each learning condition, shown separately for base and instruction-tuned models. Bars show AND minus OR accuracy, so positive coefficients favor conjunctive definitions and negative coefficients favor disjunctive definitions. The zero line marks no estimated difference. GPT-5.4 is excluded from these analyses.}
  \label{fig:lexical_effect_representation}
\end{figure*}

Lexical Category Inference includes an additional manipulation that is not fully captured by the aggregate task-level analysis. In the medium and hard settings, category definitions are constructed as either conjunctions or disjunctions of semantic conditions. The easy setting uses single-condition category definitions and is shared across the two logic conditions. We therefore focus on the medium and hard settings in Tables~\ref{tab:lexical-logic-by-model} and~\ref{tab:lexical-logic-by-difficulty}, and use Figure~\ref{fig:lexical_effect_representation} to summarize the corresponding AND--OR contrast for the open-weight models.

\begin{table*}[t]
  \centering
  \small
  \setlength{\tabcolsep}{6pt}
  \renewcommand{\arraystretch}{1.12}
  \begin{tabular*}{\textwidth}{@{}l@{\extracolsep{\fill}}lccc@{}}
    \toprule
    \textbf{Model group} & \textbf{Logic condition} & \textbf{Rules} & \textbf{Examples} & \textbf{Combined} \\
    \midrule
    \multirow[t]{2}{*}{Base models}
      & AND & 58.8 & 63.8 & 62.7 \\
      & OR  & 58.8 & 56.8 & 58.2 \\
    \midrule
    \multirow[t]{2}{*}{Instruction-tuned models}
      & AND & 61.5 & 60.9 & 59.8 \\
      & OR  & 59.3 & 50.2 & 62.3 \\
    \midrule
    \multirow[t]{2}{*}{GPT-5.4}
      & AND & 67.0 & 81.7 & 76.9 \\
      & OR  & 85.2 & 65.8 & 91.2 \\
    \bottomrule
    \end{tabular*}
    \caption{Lexical Category Inference accuracy by logic condition, model group, and learning condition. For base and instruction-tuned model groups, values are mean accuracies across models and the medium and hard difficulty levels. For GPT-5.4, values are averaged over the medium and hard difficulty levels. Base and instruction-tuned model groups include open-weight models only.}
      \label{tab:lexical-logic-by-model}
\end{table*}

Table~\ref{tab:lexical-logic-by-model} and Figure~\ref{fig:lexical_effect_representation} support our hypothesis for example-based learning. Under example-based learning, all three model groups perform better on AND definitions than on OR definitions. This is further confirmed by the statistical analysis showing a significant AND advantage for both base and instruction-tuned models only in example-based learning. As discussed in Section~\ref{subsubsec:lexical-category-inference}, this is consistent with the explanation that disjunctive category definitions being more difficult to be induced from examples when the primary mode of inference is similarity, since valid members of the category may not share features.

The same AND-over-OR pattern does not consistently appear in rule-based learning. In Table~\ref{tab:lexical-logic-by-model}, rule-based learning shows little difference between AND and OR for the open-weight model groups, and GPT-5.4 even performs better on OR definitions. Combined learning also does not show a stable AND advantage across model groups. Again, this is consistent with our hypothesis in Section~\ref{subsubsec:lexical-category-inference} that rules explicitly surface the underlying logical operators and therefore would not be subject to the same disadvantage in disjunctive category inferences. Table~\ref{tab:lexical-logic-by-difficulty} furthermore shows that the AND-OR gap increases with difficulty in example-based learning and not in rule-based learning.

\begin{table*}[t]
  \centering
  \small
  \setlength{\tabcolsep}{5pt}
  \renewcommand{\arraystretch}{1.12}
  \begin{tabular*}{\textwidth}{@{}l@{\extracolsep{\fill}}llccc@{}}
    \toprule
    \textbf{Model group} & \textbf{Difficulty} & \textbf{Logic condition} & \textbf{Rules} & \textbf{Examples} & \textbf{Combined} \\
    \midrule
    \multirow[t]{4}{*}{Base models}
      & Medium & AND & 59.8 & 64.7 & 62.1 \\
      & Medium & OR  & 59.0 & 58.1 & 58.4 \\
      & Hard   & AND & 57.9 & 63.0 & 63.2 \\
      & Hard   & OR  & 58.5 & 55.5 & 57.9 \\
    \midrule
    \multirow[t]{4}{*}{Instruction-tuned models}
      & Medium & AND & 62.0 & 60.4 & 59.0 \\
      & Medium & OR  & 57.9 & 53.9 & 62.7 \\
      & Hard   & AND & 61.1 & 61.3 & 60.6 \\
      & Hard   & OR  & 60.7 & 46.5 & 61.9 \\
    \midrule
    \multirow[t]{4}{*}{GPT-5.4}
      & Medium & AND & 66.3 & 80.3 & 79.2 \\
      & Medium & OR  & 85.0 & 67.3 & 90.5 \\
      & Hard   & AND & 67.7 & 83.0 & 74.7 \\
      & Hard   & OR  & 85.3 & 64.2 & 91.8 \\
    \bottomrule
   \end{tabular*}
    \caption{Lexical Category Inference accuracy by difficulty, logic condition, model group, and learning condition. For base and instruction-tuned model groups, values are mean accuracies averaged over models. Base and instruction-tuned model groups include open-weight models only. The easy setting is omitted because it uses the same single-condition category definitions in both logic conditions.}
  \label{tab:lexical-logic-by-difficulty}
\end{table*}

\clearpage

\section{Additional Statistical Analyses}
\label{sec:appendix-additional-stats}

\begin{figure*}[t]
  \includegraphics[width=2\columnwidth]{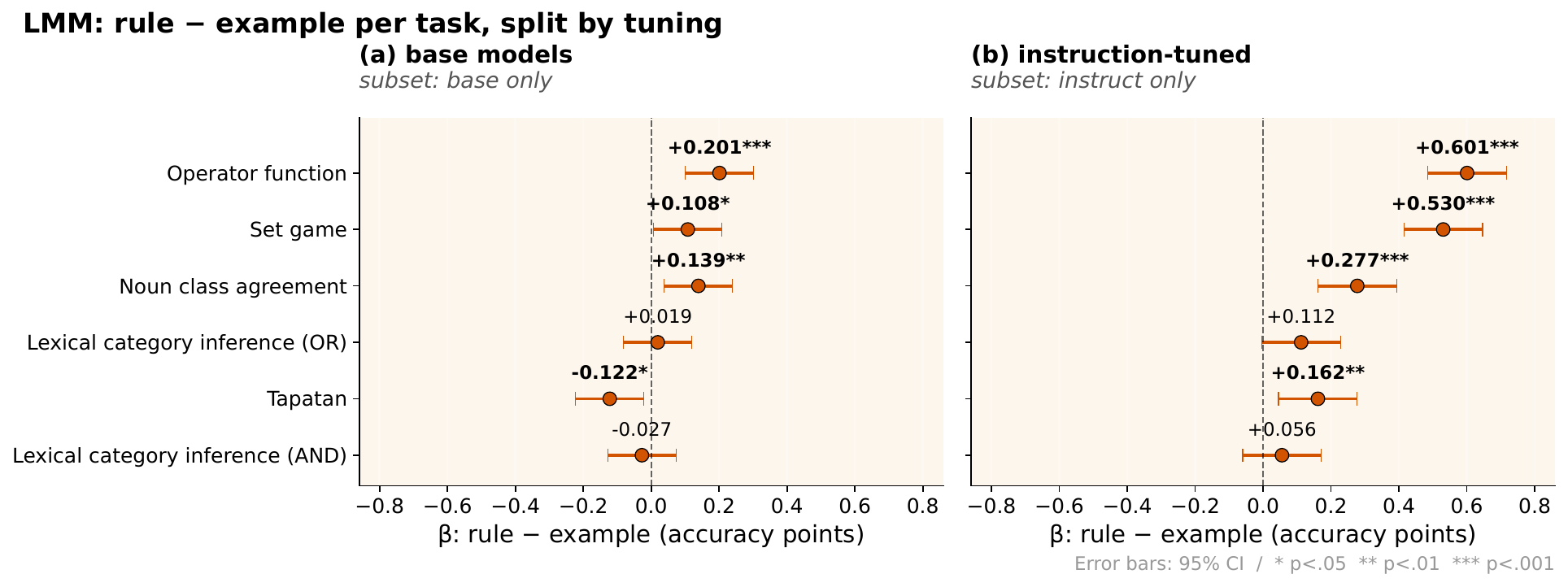}
  \caption{Task-specific rules-minus-examples contrasts, shown separately for base and instruction-tuned models. Positive coefficients indicate higher accuracy with rules, while negative coefficients indicate higher accuracy with examples. Points show estimated \(\beta\) coefficients on the zero-to-one accuracy scale, horizontal bars show 95\% confidence intervals, and the dashed zero line marks no estimated difference. Significance markers denote \(^{*}p<.05\), \(^{**}p<.01\), and \(^{***}p<.001\). GPT-5.4 is excluded from these analyses.}
  \label{fig:by-task-re-tuning}
\end{figure*}

\begin{figure*}[t]
  \includegraphics[width=2\columnwidth]{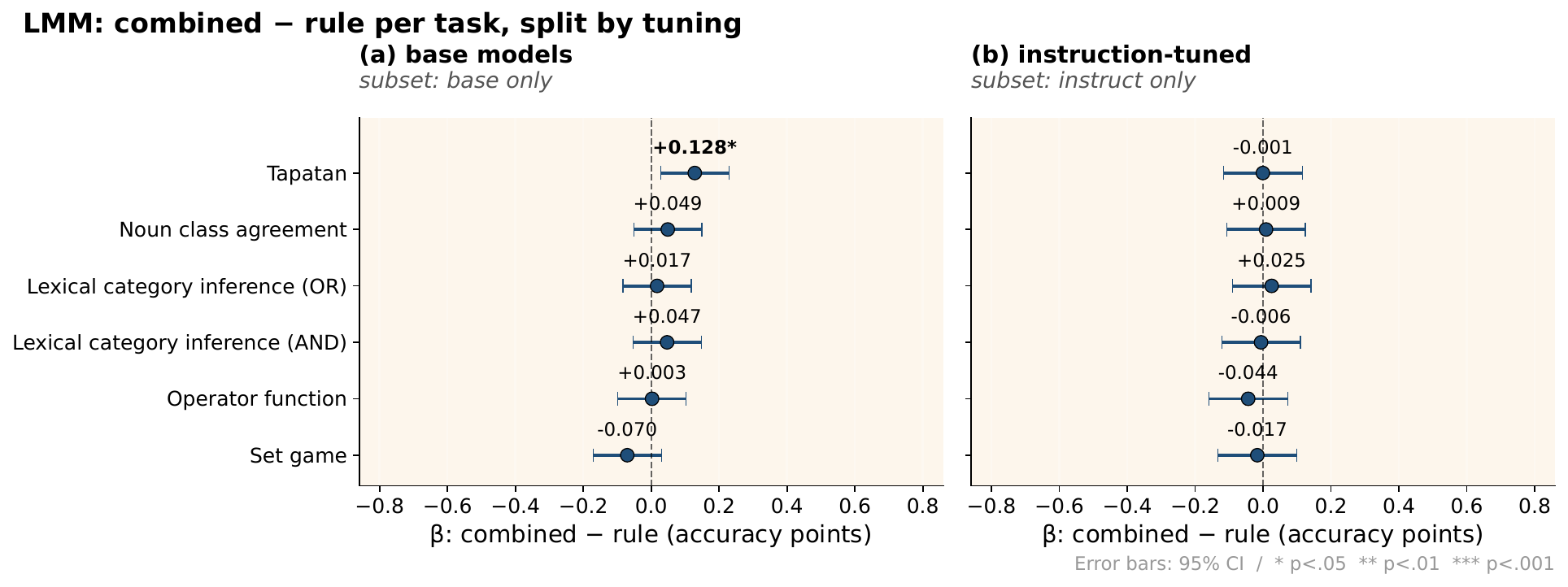}
  \caption{Task-specific combined-minus-rules contrasts, shown separately for base and instruction-tuned models. Positive coefficients indicate higher accuracy with combined prompting, while negative coefficients indicate higher accuracy with rules alone. Points show estimated \(\beta\) coefficients on the zero-to-one accuracy scale, horizontal bars show 95\% confidence intervals, and the dashed zero line marks no estimated difference. Significance markers denote \(^{*}p<.05\), \(^{**}p<.01\), and \(^{***}p<.001\). GPT-5.4 is excluded from these analyses.}
  \label{fig:by-task-cr-tuning}
\end{figure*}

Figures~\ref{fig:by-task-re-tuning} and \ref{fig:by-task-cr-tuning} provide additional statistical analyses that split the by-task results in Figure~\ref{fig:rule-example-combined-task-mod} by tuning status (base vs. instruction-tuned).

\section{Full Prompts}
\label{sec:appendix-rules-prompts}

We provide the full prompts used for our learning experiments. Figures~\ref{fig:prompt-set-game-easy} through \ref{fig:prompt-lexical-category-inference-and-hard} show the prompts for rule-based learning. The prompt structures for the example-based and combined learning settings are shown in Table~\ref{tab:prompt-structure}.

\begin{figure*}[p]
\centering
\begin{tcolorbox}[title=Set game - easy, width=\textwidth]
\small\ttfamily\raggedright\sloppy

\textbf{Task overview}\\
You will be shown 9 cards on the board and have to select 3 cards that form a GAME-SET based on the following rules.

\textbf{Rules}
\begin{enumerate}[leftmargin=1.5em, itemsep=1pt, topsep=2pt]
    \item Each card has two attributes, an animal and a biome.
    \item There are 10 types of animals, armadillo, axolotl, bat, cat, fox, horse, rabbit, sheep, turtle, wolf.
    \item There are 10 types of biomes, plain, savanna, desert, swamp, forest, jungle, taiga, hill, badland, tundra.
    \item A GAME-SET is a set of three cards. For each attribute, animal and biome, the three cards must be either all the same or all different. For example, if 2 of the cards have the same value and 1 has a different value, the set is not valid.
\end{enumerate}

\textbf{Input}\\
Here is the board.

\promptplaceholder{BOARD}

\textbf{Answer format}\\
Based on the rules, tell me which three cards here constitute a GAME-SET. Type the cards in the following format:

\begin{quote}
\ttfamily
\textbackslash boxed\{First card: CARD1\\
Second card: CARD2\\
Third card: CARD3\}
\end{quote}

\end{tcolorbox}
\caption{Rules-mode prompt for Set Game at easy difficulty.}
\label{fig:prompt-set-game-easy}
\end{figure*}

\begin{figure*}[p]
\centering
\begin{tcolorbox}[title=Set game - medium, width=\textwidth]
\small\ttfamily\raggedright\sloppy

\textbf{Task overview}\\
You will be shown 9 cards on the board and have to select 3 cards that form a GAME-SET based on the following rules.

\textbf{Rules}
\begin{enumerate}[leftmargin=1.5em, itemsep=1pt, topsep=2pt]
    \item Each card has three attributes, an animal, a biome, and a food.
    \item There are 10 types of animals, armadillo, axolotl, bat, cat, fox, horse, rabbit, sheep, turtle, wolf.
    \item There are 10 types of biomes, plain, savanna, desert, swamp, forest, jungle, taiga, hill, badland, tundra.
    \item There are 10 types of food, apple, potato, bread, cake, carrot, mutton, beef, cookie, pie, melon.
    \item A GAME-SET is a set of three cards. For each attribute, animal, biome, and food, the three cards must be either all the same or all different. For example, if 2 of the cards have the same value and 1 has a different value, the set is not valid.
\end{enumerate}

\textbf{Input}\\
Here is the board.

\promptplaceholder{BOARD}

\textbf{Answer format}\\
Based on the rules, tell me which three cards here constitute a GAME-SET. Type the cards in the following format:

\begin{quote}
\ttfamily
\textbackslash boxed\{First card: CARD1\\
Second card: CARD2\\
Third card: CARD3\}
\end{quote}

\end{tcolorbox}
\caption{Rules-mode prompt for Set Game at medium difficulty.}
\label{fig:prompt-set-game-medium}
\end{figure*}

\begin{figure*}[p]
\centering
\begin{tcolorbox}[title=Set game - hard, width=\textwidth]
\small\ttfamily\raggedright\sloppy

\textbf{Task overview}\\
You will be shown 9 cards on the board and have to select 3 cards that form a GAME-SET based on the following rules.

\textbf{Rules}
\begin{enumerate}[leftmargin=1.5em, itemsep=1pt, topsep=2pt]
    \item Each card has a number and three attributes, an animal, a biome, and a food.
    \item Numbers range from 1 to 12.
    \item There are 10 types of animals, armadillo, axolotl, bat, cat, fox, horse, rabbit, sheep, turtle, wolf.
    \item There are 10 types of biomes, plain, savanna, desert, swamp, forest, jungle, taiga, hill, badland, tundra.
    \item There are 10 types of food, apple, potato, bread, cake, carrot, mutton, beef, cookie, pie, melon.
    \item A GAME-SET is a set of three cards. For each attribute, animal, biome, and food, the three cards must be either all the same or all different. For example, if 2 of the cards have the same value and 1 has a different value, the set is not valid.
    \item For the number attribute only, 2 of the cards should have the same number and 1 should have a different number in order for the set to be valid.
\end{enumerate}

\textbf{Input}\\
Here is the board.

\promptplaceholder{BOARD}

\textbf{Answer format}\\
Based on the rules, tell me which three cards here constitute a GAME-SET. Type the cards in the following format:

\begin{quote}
\ttfamily
\textbackslash boxed\{First card: CARD1\\
Second card: CARD2\\
Third card: CARD3\}
\end{quote}

\end{tcolorbox}
\caption{Rules-mode prompt for Set Game at hard difficulty.}
\label{fig:prompt-set-game-hard}
\end{figure*}

\begin{figure*}[p]
\centering
\begin{tcolorbox}[title=Tapatan - easy, width=\textwidth]
\small\ttfamily\raggedright\sloppy

\textbf{Task overview}\\
You will be learning a new board game. You will then be given one final board state from that game. Determine the current game outcome.

\textbf{Rules}
\begin{enumerate}[leftmargin=1.5em, itemsep=1pt, topsep=2pt]
    \item The game is played on a \(3{\times}3\) board.
    \item The input board has 3 rows and 3 columns. Each cell contains \texttt{A}, \texttt{B}, or \texttt{.}. \texttt{A} means the cell contains a piece from Player \(A\), \texttt{B} means the cell contains a piece from Player \(B\), and \texttt{.} means the cell is empty.
    \item Rows are shown from top to bottom, and columns are shown from left to right. The cell in column \(x\) and row \(y\) has coordinate \((x,y)\), where \(0 \leq x,y < 3\).
    \item Each player can have at most 3 pieces on the board.
    \item Orthogonal neighbors are always adjacent. A diagonal pair of neighboring cells is adjacent only when those cells have even parity, meaning \(x+y\) is even for the cells in that diagonal pair.
    \item A player wins if that player has 3 pieces in a straight line, horizontal, vertical, or diagonal, and each consecutive pair of cells in that line is adjacent under rule 5.
    \item The board states in this task contain at most one winning player. If \(A\) has a winning line, output \texttt{A win}. If \(B\) has a winning line, output \texttt{B win}. If neither player has a winning line, output \texttt{continue}.
    \item You must output a label that is exactly one of \texttt{A win}, \texttt{B win}, or \texttt{continue}.
\end{enumerate}

\textbf{Input}\\
\texttt{BOARD}\\
\promptplaceholder{BOARD}

\textbf{Answer}\\
\texttt{LABEL:}

\end{tcolorbox}
\caption{Rules-mode prompt for Tapatan at easy difficulty.}
\label{fig:prompt-tapatan-easy}
\end{figure*}

\begin{figure*}[p]
\centering
\begin{tcolorbox}[title=Tapatan - medium, width=\textwidth]
\small\ttfamily\raggedright\sloppy

\textbf{Task overview}\\
You will be learning a new board game. You will then be given one final board state from that game. Determine the current game outcome.

\textbf{Rules}
\begin{enumerate}[leftmargin=1.5em, itemsep=1pt, topsep=2pt]
    \item The game is played on a \(5{\times}5\) board.
    \item The input board has 5 rows and 5 columns. Each cell contains \texttt{A}, \texttt{B}, or \texttt{.}. \texttt{A} means the cell contains a piece from Player \(A\), \texttt{B} means the cell contains a piece from Player \(B\), and \texttt{.} means the cell is empty.
    \item Rows are shown from top to bottom, and columns are shown from left to right. The cell in column \(x\) and row \(y\) has coordinate \((x,y)\), where \(0 \leq x,y < 5\).
    \item Each player can have at most 4 pieces on the board.
    \item Orthogonal neighbors are always adjacent. A diagonal pair of neighboring cells is adjacent only when those cells have even parity, meaning \(x+y\) is even for the cells in that diagonal pair.
    \item A player wins if that player has 4 pieces in a straight line, horizontal, vertical, or diagonal, and each consecutive pair of cells in that line is adjacent under rule 5.
    \item The board states in this task contain at most one winning player. If \(A\) has a winning line, output \texttt{A win}. If \(B\) has a winning line, output \texttt{B win}. If neither player has a winning line, output \texttt{continue}.
    \item You must output a label that is exactly one of \texttt{A win}, \texttt{B win}, or \texttt{continue}.
\end{enumerate}

\textbf{Input}\\
\texttt{BOARD}\\
\promptplaceholder{BOARD}

\textbf{Answer}\\
\texttt{LABEL:}

\end{tcolorbox}
\caption{Rules-mode prompt for Tapatan at medium difficulty.}
\label{fig:prompt-tapatan-medium}
\end{figure*}

\begin{figure*}[p]
\centering
\begin{tcolorbox}[title=Tapatan - hard, width=\textwidth]
\small\ttfamily\raggedright\sloppy

\textbf{Task overview}\\
You will be learning a new board game. You will then be given one final board state from that game. Determine the current game outcome.

\textbf{Rules}
\begin{enumerate}[leftmargin=1.5em, itemsep=1pt, topsep=2pt]
    \item The game is played on a \(7{\times}7\) board.
    \item The input board has 7 rows and 7 columns. Each cell contains \texttt{A}, \texttt{B}, or \texttt{.}. \texttt{A} means the cell contains a piece from Player \(A\), \texttt{B} means the cell contains a piece from Player \(B\), and \texttt{.} means the cell is empty.
    \item Rows are shown from top to bottom, and columns are shown from left to right. The cell in column \(x\) and row \(y\) has coordinate \((x,y)\), where \(0 \leq x,y < 7\).
    \item Each player can have at most 5 pieces on the board.
    \item Orthogonal neighbors are always adjacent. A diagonal pair of neighboring cells is adjacent only when those cells have even parity, meaning \(x+y\) is even for the cells in that diagonal pair.
    \item A player wins if that player has 5 pieces in a straight line, horizontal, vertical, or diagonal, and each consecutive pair of cells in that line is adjacent under rule 5.
    \item The board states in this task contain at most one winning player. If \(A\) has a winning line, output \texttt{A win}. If \(B\) has a winning line, output \texttt{B win}. If neither player has a winning line, output \texttt{continue}.
    \item You must output a label that is exactly one of \texttt{A win}, \texttt{B win}, or \texttt{continue}.
\end{enumerate}

\textbf{Input}\\
\texttt{BOARD}\\
\promptplaceholder{BOARD}

\textbf{Answer}\\
\texttt{LABEL:}

\end{tcolorbox}
\caption{Rules-mode prompt for Tapatan at hard difficulty.}
\label{fig:prompt-tapatan-hard}
\end{figure*}

\begin{figure*}[p]
\centering
\begin{tcolorbox}[title=Operator function - easy, width=\textwidth]
\small\ttfamily\raggedright\sloppy

\textbf{Task overview}\\
You will be shown a problem that uses a newly defined operator function based on the following rules and need to calculate the answer for the problem.

\textbf{Rules}
\begin{enumerate}[leftmargin=1.5em, itemsep=1pt, topsep=2pt]
    \item The new operator function f is composed of two arithmetic operators, and its inputs are integers.
    \item Let x, y, z be inputs of the function.
    \item To compute f(x,y,z), first multiply x and y.
    \item Then, add z to the result from the previous operation.
\end{enumerate}

\textbf{Input}\\
Problem: \texttt{f(}\promptplaceholder{x}\texttt{,}\promptplaceholder{y}\texttt{,}\promptplaceholder{z}\texttt{)}

\textbf{Answer format}\\
Answer(place the result in \texttt{\textbackslash boxed\{\}}): \texttt{\textbackslash boxed\{\}}

\end{tcolorbox}
\caption{Rules-mode prompt for Operator Function at easy difficulty.}
\label{fig:prompt-operator-function-easy}
\end{figure*}

\begin{figure*}[p]
\centering
\begin{tcolorbox}[title=Operator function - medium, width=\textwidth]
\small\ttfamily\raggedright\sloppy

\textbf{Task overview}\\
You will be shown a problem that uses a newly defined operator function based on the following rules and need to calculate the answer for the problem.

\textbf{Rules}
\begin{enumerate}[leftmargin=1.5em, itemsep=1pt, topsep=2pt]
    \item The new operator function g is composed of three arithmetic operators, and its inputs are integers.
    \item Let x, y, z, a be inputs of the function.
    \item To compute g(x,y,z,a), first subtract y from x.
    \item Second, multiply the result from the previous operation by z.
    \item Last, adds a to the result of the previous operation. 
\end{enumerate}

\textbf{Input}\\
Problem: \texttt{g(}\promptplaceholder{x}\texttt{,}\promptplaceholder{y}\texttt{,}\promptplaceholder{z}\texttt{,}\promptplaceholder{a}\texttt{)}

\textbf{Answer format}\\
Answer (place the result in \texttt{\textbackslash boxed\{\}}): \texttt{\textbackslash boxed\{\}}

\end{tcolorbox}
\caption{Rules-mode prompt for Operator Function at medium difficulty.}
\label{fig:prompt-operator-function-medium}
\end{figure*}

\begin{figure*}[p]
\centering
\begin{tcolorbox}[title=Operator function - hard, width=\textwidth]
\small\ttfamily\raggedright\sloppy

\textbf{Task overview}\\
You will be shown a problem that uses a newly defined operator function based on the following rules and need to calculate the answer for the problem.

\textbf{Rules}
\begin{enumerate}[leftmargin=1.5em, itemsep=1pt, topsep=2pt]
    \item The new operator function h is composed of four arithmetic operators, and its inputs are integers.
    \item Let x, y, z, a, b be inputs of the function.
    \item To compute h(x,y,z,a,b), first add x and y.
    \item Second, multiply the result from the previous operation by z.
    \item Third, subtract a from the result of the previous operation.
    \item Last, take the remainder when the result is divided by b.
\end{enumerate}

\textbf{Input}\\
Problem: \texttt{h(}\promptplaceholder{x}\texttt{,}\promptplaceholder{y}\texttt{,}\promptplaceholder{z}\texttt{,}\promptplaceholder{a}\texttt{,}\promptplaceholder{b}\texttt{)}

\textbf{Answer format}\\
Answer(place the result in \texttt{\textbackslash boxed\{\}}): \texttt{\textbackslash boxed\{\}}

\end{tcolorbox}
\caption{Rules-mode prompt for Operator Function at hard difficulty.}
\label{fig:prompt-operator-function-hard}
\end{figure*}

\begin{figure*}[p]
\centering
\begin{tcolorbox}[title=Noun class agreement - easy, width=\textwidth]
\small\ttfamily\raggedright\sloppy

\textbf{Task overview}\\
You will be given one sentence in a synthetic language. Decide whether the sentence is grammatical.

\textbf{Rules}
\begin{enumerate}[leftmargin=1.5em, itemsep=1pt, topsep=2pt]
    \item There are 2 noun classes, \(A\) and \(B\).
    \item Each noun phrase has the form \texttt{DET NOUN}.
    \item The class \(A\) determiner is \texttt{ka}, and the class \(B\) determiner is \texttt{ti}.
    \item The class \(A\) nouns are \texttt{tac}, \texttt{sular}, \texttt{fep}, \texttt{wug}. The class \(B\) nouns are \texttt{bim}, \texttt{noko}, \texttt{glarn}, \texttt{zesh}.
    \item Sentences follow the template \texttt{DET NOUN VERB DET NOUN}.
    \item Verbs are plain roots like \texttt{dax}, \texttt{miv}, \texttt{lorp} and never change.
    \item A sentence is grammatical if and only if, in each noun phrase, the determiner matches the noun's class.
    \item Your answer must be exactly one word, Yes or No.
\end{enumerate}

\textbf{Input}\\
Sentence: \promptplaceholder{SENTENCE}

\textbf{Answer}\\
Answer:

\end{tcolorbox}
\caption{Rules-mode prompt for Noun Class Agreement at easy difficulty.}
\label{fig:prompt-noun-class-agreement-easy}
\end{figure*}

\begin{figure*}[p]
\centering
\begin{tcolorbox}[title=Noun class agreement - medium, width=\textwidth]
\small\ttfamily\raggedright\sloppy

\textbf{Task overview}\\
You will be given one sentence in a synthetic language. Decide whether the sentence is grammatical.

\textbf{Rules}
\begin{enumerate}[leftmargin=1.5em, itemsep=1pt, topsep=2pt]
    \item There are 4 noun classes, \(A\), \(B\), \(C\), and \(D\).
    \item A noun phrase has the form \texttt{DET ADJ-SUFF NOUN}.
    \item The class \(A\) determiner is \texttt{ka}, the class \(B\) determiner is \texttt{ti}, the class \(C\) determiner is \texttt{su}, and the class \(D\) determiner is \texttt{vo}.
    \item Adjectives have the form \texttt{STEM-SUFFIX}. The suffix must match the noun's class.
    \item The class \(A\) adjective suffix is \texttt{-en}, the class \(B\) adjective suffix is \texttt{-os}, the class \(C\) adjective suffix is \texttt{-im}, and the class \(D\) adjective suffix is \texttt{-at}.
    \item Example adjective tokens are \texttt{glim-en}, \texttt{prun-os}, \texttt{fen-im}, \texttt{zay-at}.
    \item The class \(A\) nouns are \texttt{tav}, \texttt{fap}. The class \(B\) nouns are \texttt{bim}, \texttt{glarn}. The class \(C\) nouns are \texttt{sular}, \texttt{wug}. The class \(D\) nouns are \texttt{noko}, \texttt{zesh}.
    \item Sentences follow the template \texttt{DET ADJ-SUFF NOUN VERB DET ADJ-SUFF NOUN}.
    \item Verbs are plain roots like \texttt{dax}, \texttt{miv}, \texttt{lorp} and never change.
    \item A sentence is grammatical if and only if, in each noun phrase, both the determiner and the adjective suffix match the noun's class.
    \item Your answer must be exactly one word, Yes or No.
\end{enumerate}

\textbf{Input}\\
Sentence: \promptplaceholder{SENTENCE}

\textbf{Answer}\\
Answer:

\end{tcolorbox}
\caption{Rules-mode prompt for Noun Class Agreement at medium difficulty.}
\label{fig:prompt-noun-class-agreement-medium}
\end{figure*}

\begin{figure*}[p]
\centering
\begin{tcolorbox}[title=Noun class agreement - hard, width=\textwidth]
\small\ttfamily\raggedright\sloppy

\textbf{Task overview}\\
You will be given one sentence in a synthetic language. Decide whether the sentence is grammatical.

\textbf{Rules}
\begin{enumerate}[leftmargin=1.5em, itemsep=1pt, topsep=2pt]
    \item There are 6 noun classes, \(A\), \(B\), \(C\), \(D\), \(E\), and \(F\).
    \item A noun phrase has the form \texttt{DET ADJ-SUFF NOUN}.
    \item The class \(A\) determiner is \texttt{ka}, the class \(B\) determiner is \texttt{ti}, the class \(C\) determiner is \texttt{su}, the class \(D\) determiner is \texttt{vo}, the class \(E\) determiner is \texttt{ne}, and the class \(F\) determiner is \texttt{la}.
    \item Adjectives have the form \texttt{STEM-SUFFIX}. The suffix must match the noun's class.
    \item The class \(A\) adjective suffix is \texttt{-en}, the class \(B\) adjective suffix is \texttt{-os}, the class \(C\) adjective suffix is \texttt{-im}, the class \(D\) adjective suffix is \texttt{-at}, the class \(E\) adjective suffix is \texttt{-uk}, and the class \(F\) adjective suffix is \texttt{-esh}.
    \item The verb has the form \texttt{PREFIX-ROOT-SUFFIX}.
    \item The prefix must match the class of the subject noun, and the suffix must match the class of the object noun.
    \item The class \(A\) subject prefix is \texttt{ge-}, the class \(B\) subject prefix is \texttt{du-}, the class \(C\) subject prefix is \texttt{ri-}, the class \(D\) subject prefix is \texttt{zo-}, the class \(E\) subject prefix is \texttt{pa-}, and the class \(F\) subject prefix is \texttt{li-}.
    \item The class \(A\) object suffix is \texttt{-an}, the class \(B\) object suffix is \texttt{-eb}, the class \(C\) object suffix is \texttt{-ig}, the class \(D\) object suffix is \texttt{-ot}, the class \(E\) object suffix is \texttt{-ul}, and the class \(F\) object suffix is \texttt{-er}.
    \item Example verb tokens are \texttt{ge-dax-eb}, \texttt{pa-miv-ig}, \texttt{du-lorp-er}.
    \item The class \(A\) nouns are \texttt{tav}, \texttt{fap}. The class \(B\) nouns are \texttt{bim}, \texttt{glarn}. The class \(C\) nouns are \texttt{sular}, \texttt{wug}. The class \(D\) nouns are \texttt{noko}, \texttt{zesh}. The class \(E\) nouns are \texttt{tac}, \texttt{fep}. The class \(F\) nouns are \texttt{lurn}, \texttt{prax}.
    \item Sentences follow the template \texttt{DET ADJ-SUFF NOUN PREFIX-ROOT-SUFFIX DET ADJ-SUFF NOUN}.
    \item Verb roots are plain roots like \texttt{dax}, \texttt{miv}, \texttt{lorp}. Only the verb's prefix and suffix change.
    \item A sentence is grammatical if and only if the subject noun phrase's determiner and adjective suffix match the subject noun's class, the object noun phrase's determiner and adjective suffix match the object noun's class, the verb prefix matches the subject noun's class, and the verb suffix matches the object noun's class.
    \item Your answer must be exactly one word, Yes or No.
\end{enumerate}

\textbf{Input}\\
Sentence: \promptplaceholder{SENTENCE}

\textbf{Answer}\\
Answer:

\end{tcolorbox}
\caption{Rules-mode prompt for Noun Class Agreement at hard difficulty.}
\label{fig:prompt-noun-class-agreement-hard}
\end{figure*}

\begin{figure*}[p]
\centering
\begin{tcolorbox}[title=Lexical category inference - easy, width=\textwidth]
\small\ttfamily\raggedright\sloppy

\textbf{Task overview}\\
You will be shown a list of four words. Decide whether this list is correct based on the following rules.

\textbf{Rules}
\begin{enumerate}[leftmargin=1.5em, itemsep=1pt, topsep=2pt]
    \item There are exactly four categories, Category 1, Category 2, Category 3, and Category 4.
    \item Each category has the meaning shown in the category definitions below.
    \item A correct list contains exactly four comma-separated items.
    \item In a correct list, the first item is a member of Category 1, the second item is a member of Category 2, and so on.
    \item If a category says ``at least one'', an item may satisfy any listed description. If it says ``all'', an item must satisfy every listed description.
    \item Your answer must be exactly one word, Yes or No.
\end{enumerate}

\textbf{Category definitions}\\
Category 1: \promptplaceholder{single-condition definition for Category 1}\\
Category 2: \promptplaceholder{single-condition definition for Category 2}\\
Category 3: \promptplaceholder{single-condition definition for Category 3}\\
Category 4: \promptplaceholder{single-condition definition for Category 4}

\textbf{Input}\\
List: \promptplaceholder{word 1}, \promptplaceholder{word 2}, \promptplaceholder{word 3}, \promptplaceholder{word 4}

\textbf{Answer}\\
Answer:

\end{tcolorbox}
\caption{Rules-mode prompt for Lexical Category Inference at easy difficulty.}
\label{fig:prompt-lexical-category-inference-easy}
\end{figure*}

\begin{figure*}[p]
\centering
\begin{tcolorbox}[title=Lexical category inference: OR - medium, width=\textwidth]
\small\ttfamily\raggedright\sloppy

\textbf{Task overview}\\
You will be shown a list of four words. Decide whether this list is correct based on the following rules.

\textbf{Rules}
\begin{enumerate}[leftmargin=1.5em, itemsep=1pt, topsep=2pt]
    \item There are exactly four categories, Category 1, Category 2, Category 3, and Category 4.
    \item Each category has the meaning shown in the category definitions below.
    \item A correct list contains exactly four comma-separated items.
    \item In a correct list, the first item is a member of Category 1, the second item is a member of Category 2, and so on.
    \item If a category says ``at least one'', an item may satisfy any listed description. If it says ``all'', an item must satisfy every listed description.
    \item Your answer must be exactly one word, Yes or No.
\end{enumerate}

\textbf{Category definitions}\\
Category 1: items that fit at least one of these descriptions, \promptplaceholder{condition 1A}; \promptplaceholder{condition 1B}\\
Category 2: items that fit at least one of these descriptions, \promptplaceholder{condition 2A}; \promptplaceholder{condition 2B}\\
Category 3: items that fit at least one of these descriptions, \promptplaceholder{condition 3A}; \promptplaceholder{condition 3B}\\
Category 4: items that fit at least one of these descriptions, \promptplaceholder{condition 4A}; \promptplaceholder{condition 4B}

\textbf{Input}\\
List: \promptplaceholder{word 1}, \promptplaceholder{word 2}, \promptplaceholder{word 3}, \promptplaceholder{word 4}

\textbf{Answer}\\
Answer:

\end{tcolorbox}
\caption{Rules-mode prompt for Lexical Category Inference at medium difficulty with disjunctive categories.}
\label{fig:prompt-lexical-category-inference-or-medium}
\end{figure*}

\begin{figure*}[p]
\centering
\begin{tcolorbox}[title=Lexical category inference: AND - medium, width=\textwidth]
\small\ttfamily\raggedright\sloppy

\textbf{Task overview}\\
You will be shown a list of four words. Decide whether this list is correct based on the following rules.

\textbf{Rules}
\begin{enumerate}[leftmargin=1.5em, itemsep=1pt, topsep=2pt]
    \item There are exactly four categories, Category 1, Category 2, Category 3, and Category 4.
    \item Each category has the meaning shown in the category definitions below.
    \item A correct list contains exactly four comma-separated items.
    \item In a correct list, the first item is a member of Category 1, the second item is a member of Category 2, and so on.
    \item If a category says ``at least one'', an item may satisfy any listed description. If it says ``all'', an item must satisfy every listed description.
    \item Your answer must be exactly one word, Yes or No.
\end{enumerate}

\textbf{Category definitions}\\
Category 1: items that fit all of these descriptions, \promptplaceholder{condition 1A}; \promptplaceholder{condition 1B}\\
Category 2: items that fit all of these descriptions, \promptplaceholder{condition 2A}; \promptplaceholder{condition 2B}\\
Category 3: items that fit all of these descriptions, \promptplaceholder{condition 3A}; \promptplaceholder{condition 3B}\\
Category 4: items that fit all of these descriptions, \promptplaceholder{condition 4A}; \promptplaceholder{condition 4B}

\textbf{Input}\\
List: \promptplaceholder{word 1}, \promptplaceholder{word 2}, \promptplaceholder{word 3}, \promptplaceholder{word 4}

\textbf{Answer}\\
Answer:
\end{tcolorbox}
\caption{Rules-mode prompt for Lexical Category Inference at medium difficulty with conjunctive categories.}
\label{fig:prompt-lexical-category-inference-and-medium}
\end{figure*}

\begin{figure*}[p]
\centering
\begin{tcolorbox}[title=Lexical category inference: OR - hard, width=\textwidth]
\small\ttfamily\raggedright\sloppy

\textbf{Task overview}\\
You will be shown a list of four words. Decide whether this list is correct based on the following rules.

\textbf{Rules}
\begin{enumerate}[leftmargin=1.5em, itemsep=1pt, topsep=2pt]
    \item There are exactly four categories, Category 1, Category 2, Category 3, and Category 4.
    \item Each category has the meaning shown in the category definitions below.
    \item A correct list contains exactly four comma-separated items.
    \item In a correct list, the first item is a member of Category 1, the second item is a member of Category 2, and so on.
    \item If a category says ``at least one'', an item may satisfy any listed description. If it says ``all'', an item must satisfy every listed description.
    \item Your answer must be exactly one word, Yes or No.
\end{enumerate}

\textbf{Category definitions}\\
Category 1: items that fit at least one of these descriptions, \promptplaceholder{condition 1A}; \promptplaceholder{condition 1B}; \promptplaceholder{condition 1C}\\
Category 2: items that fit at least one of these descriptions, \promptplaceholder{condition 2A}; \promptplaceholder{condition 2B}; \promptplaceholder{condition 2C}\\
Category 3: items that fit at least one of these descriptions, \promptplaceholder{condition 3A}; \promptplaceholder{condition 3B}; \promptplaceholder{condition 3C}\\
Category 4: items that fit at least one of these descriptions, \promptplaceholder{condition 4A}; \promptplaceholder{condition 4B}; \promptplaceholder{condition 4C}

\textbf{Input}\\
List: \promptplaceholder{word 1}, \promptplaceholder{word 2}, \promptplaceholder{word 3}, \promptplaceholder{word 4}

\textbf{Answer}\\
Answer:

\end{tcolorbox}
\caption{Rules-mode prompt for Lexical Category Inference at hard difficulty with disjunctive categories.}
\label{fig:prompt-lexical-category-inference-or-hard}
\end{figure*}

\begin{figure*}[p]
\centering
\begin{tcolorbox}[title=Lexical category inference: AND - hard, width=\textwidth]
\small\ttfamily\raggedright\sloppy
\textbf{Task overview} \\
You will be shown a list of four words. Decide whether this list is correct based on the following rules.\\

\textbf{Rules} \\
\begin{enumerate}[leftmargin=1.5em, itemsep=1pt, topsep=2pt]
    \item There are exactly four categories, Category 1, Category 2, Category 3, and Category 4.
    \item Each category has the meaning shown in the category definitions below.
    \item A correct list contains exactly four comma-separated items.
    \item In a correct list, the first item is a member of Category 1, the second item is a member of Category 2, and so on.
    \item If a category says ``at least one'', an item may satisfy any listed description. If it says ``all'', an item must satisfy every listed description.
    \item Your answer must be exactly one word, Yes or No.
\end{enumerate}

\textbf{Category definitions}\\
Category 1: items that fit all three of these descriptions, \promptplaceholder{condition 1A}; \promptplaceholder{condition 1B}; \promptplaceholder{condition 1C}\\
Category 2: items that fit all three of these descriptions, \promptplaceholder{condition 2A}; \promptplaceholder{condition 2B}; \promptplaceholder{condition 2C}\\
Category 3: items that fit all three of these descriptions, \promptplaceholder{condition 3A}; \promptplaceholder{condition 3B}; \promptplaceholder{condition 3C}\\
Category 4: items that fit all three of these descriptions, \promptplaceholder{condition 4A}; \promptplaceholder{condition 4B}; \promptplaceholder{condition 4C}

\textbf{Input}\\
List: \promptplaceholder{word 1}, \promptplaceholder{word 2}, \promptplaceholder{word 3}, \promptplaceholder{word 4}

\textbf{Answer} \\
Answer:
\end{tcolorbox}
\caption{Rules-mode prompt for Lexical Category Inference at hard difficulty with conjunctive categories.}
\label{fig:prompt-lexical-category-inference-and-hard}
\end{figure*}

\clearpage

\begin{table*}[!t]
  \centering
  \footnotesize
  \setlength{\tabcolsep}{3pt}
  \renewcommand{\arraystretch}{1.12}
  \begin{tabularx}{\textwidth}{@{}Y Y Y@{}}
    \toprule
    \textbf{Rules} & \textbf{Examples} & \textbf{Combined}  \\
    \midrule

        \textbf{Header} \newline
        \texttt{Task overview} \newline You will...\newline
        \newline
        \textbf{Rules} \newline
        \texttt{Rules}: \newline
        1. ...\newline
        \newline
        \textbf{Problem} \newline
        Input: \promptplaceholder{Input} \newline
        Output:
    &
    \textbf{Examples} \newline
    \promptplaceholder{Input} $\rightarrow$ \promptplaceholder{Output} \newline
    ...\newline
    \promptplaceholder{Input} $\rightarrow$ \promptplaceholder{Output} \newline
    
    \textbf{Problem}\newline
    \promptplaceholder{Input} $\rightarrow$
    &
    \textbf{Header} \newline
    \texttt{Task overview} \newline You will...\newline
    \newline
    \textbf{Rules} \newline
    \texttt{Rules}: \newline
    1. ...\newline
    
    \textbf{Examples} \newline
    \promptplaceholder{Input} $\rightarrow$ \promptplaceholder{Output} \newline
    ...\newline
    \promptplaceholder{Input} $\rightarrow$ \promptplaceholder{Output} \newline
    
    \textbf{Problem} \newline
    Input: \promptplaceholder{Input} \newline
    Output:
    \\

    \bottomrule
  \end{tabularx}
  \caption{Prompt structure for each learning condition. In the combined condition, examples are inserted after the rule prompt and before the test problem.}
  \label{tab:prompt-structure}
\end{table*}

\clearpage

\end{document}